\pdfoutput=1 %
\documentclass{article} %
\usepackage{iclr2027_conference,times}

\usepackage{amsmath,amsfonts,bm}

\def\eqref#1{equation~\ref{#1}}

\def\1{\bm{1}}

\DeclareMathAlphabet{\mathsfit}{\encodingdefault}{\sfdefault}{m}{sl}
\SetMathAlphabet{\mathsfit}{bold}{\encodingdefault}{\sfdefault}{bx}{n}

\usepackage{hyperref}
\usepackage{url}
\newcommand{\doi}[1]{doi: \href{https://doi.org/#1}{\begingroup\urlstyle{rm}\nolinkurl{#1}\endgroup}}
\usepackage{graphicx}
\usepackage{xcolor}
\usepackage{booktabs}
\usepackage{longtable}
\usepackage{array}
\usepackage{placeins}
\usepackage{float}



\title{ibUMAP: Coherent and Scalable\\Field Evaluation for UMAP Optimization}

\author{Bin Chen$^{1}$, Yumeng Xue$^{1}$, Patrick Paetzold$^{1}$, Yunhai Wang$^{2}$ \& Oliver Deussen$^{1}$\\
$^{1}$University of Konstanz\qquad $^{2}$Renmin University of China}

\iclrfinalcopy
\begin{document}

\maketitle
\lhead{Preprint. Under review.}

\begin{abstract}
UMAP achieves scalable layout optimization through stochastic negative sampling. However, this stochasticity can lead to unstable embeddings across reruns and downstream reuse, as the estimated repulsive forces depend on the ordering of sampling events. We present ibUMAP, a coherent field-based alternative that evaluates attraction and repulsion from a shared embedding snapshot and applies them synchronously. Its degree-weighted repulsive field is motivated by the conditional expectation of negative sampling for a fixed embedding and represented by three scalar moments, which are evaluated efficiently on CPUs and GPUs using an interpolation-based FFT scheme. This formulation avoids explicit all-pairs computations while inducing optimization dynamics that differ from those of standard online UMAP. Controlled experiments show that synchrony and kernel capping alter the local--global fidelity trade-off, whereas FFT evaluation produces small average changes in final quality. End-to-end benchmarks show median speedups of $3.29\times$ unseeded and $5.79\times$ seeded over umap-learn on CPU, and $1.44\times$ over cuML on million-scale datasets under unseeded GPU execution. These gains accompany greater run-to-run stability and measurable fidelity trade-offs.
\end{abstract}

\section{Introduction}
\label{sec:introduction}

Uniform Manifold Approximation and Projection (UMAP) is a widely used neighbor-embedding technique whose practical optimization scheme combines sampled attraction over the fuzzy graph with stochastic negative sampling in the embedding space \citep{mcinnes2020umap}. This online formulation circumvents the need to evaluate an explicit dense non-edge term, but its repeated random sampling and in-place pairwise updates make the resulting optimization trajectory sensitive to the ordering of events. Such sensitivity becomes particularly relevant when embeddings are repeatedly computed at scale or serve as intermediate representations in subsequent analysis pipelines.

\begin{figure}[t]
    \centering
    \includegraphics[width=\linewidth]{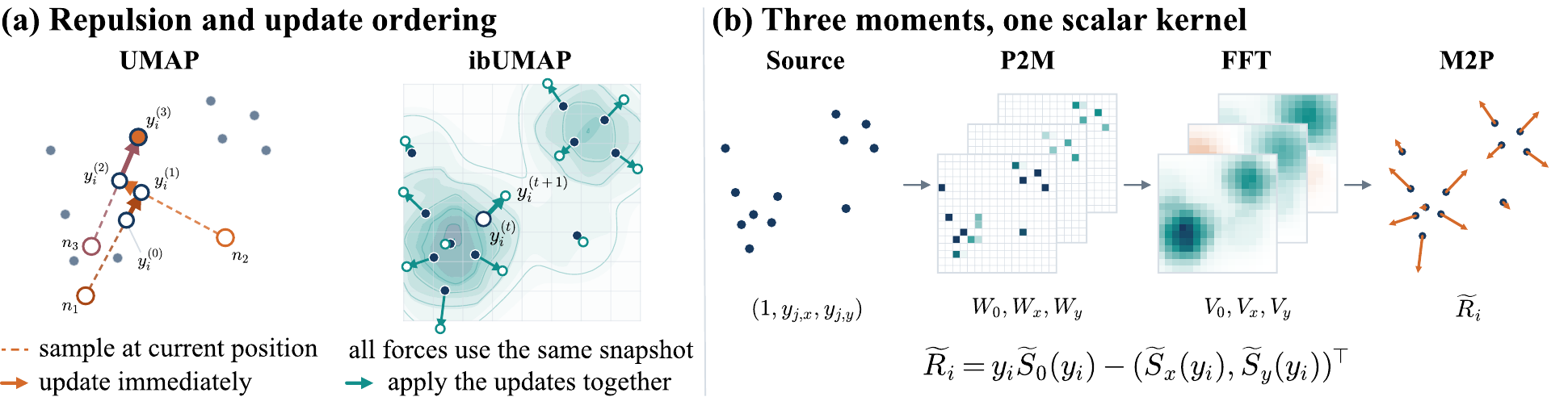}
    \caption{\textbf{ibUMAP: from sampled in-place updates to coherent field evaluation.}
    \textbf{(a)} From shared initial coordinates, UMAP negative-sampling events update the head immediately, whereas ibUMAP evaluates a radial repulsive field at one snapshot and applies updates together.
    \textbf{(b)} Particle-to-mesh (P2M) deposition, FFT convolution with a shared capped scalar kernel, and mesh-to-particle (M2P) interpolation reconstruct $\widetilde R_i$ from three scalar moments.
    }
    \label{fig:overview}
\end{figure}

The expectation underlying UMAP's negative-sampling optimizer offers a natural basis for an alternative treatment of its repulsive interactions \citep{damrich2021umap}. Drawing on earlier $N$-body formulations of nonlinear embedding gradients \citep{vladymyrov2014linear} and the interpolation-based FFT approach developed for FIt-SNE \citep{linderman2019fast}, we express degree-weighted repulsion as a continuous field that can be evaluated efficiently without explicit enumeration of all point pairs. Rather than accelerating UMAP's stochastic sampling scheme, as in existing GPU implementations \citep{nolet2021bringing}, this formulation explores an alternative computational regime with distinct optimization dynamics. ibUMAP retains UMAP's graph construction and initialization algorithms, modifying only the layout-optimization stage. Figure~\ref{fig:overview} contrasts the update mechanisms in panel~(a) and illustrates the three-moment field evaluation in panel~(b).

Replacing online sampled interactions with a synchronous field changes the optimizer as well as its computational strategy. Conclusions derived for pairwise UMAP updates \citep{islam2026shape} do not necessarily extend to synchronous force aggregation. Because informative initialization can itself affect apparent global structure \citep{kobak2021initialization}, we hold the graph and initialization fixed when studying optimizer effects. We therefore investigate two related questions: how to efficiently evaluate the degree-weighted repulsive field for a fixed embedding, and how synchronous updates, repulsion formulation, ibFFT field evaluation, and safeguards affect final embedding quality under controlled graph construction and initialization. These controlled comparisons complement end-to-end benchmarks of speed, fidelity, and repeatability.

\paragraph{Contributions.}
\emph{Formulation and implementation.} We formalize the frozen-state, degree-weighted repulsive operator under explicit negative-sampling and update conventions and derive a UMAP-specific three-moment decomposition, enabling synchronous two-dimensional FFT-based optimization on CPUs and GPUs with explicit complexity and repeatability conditions.
\emph{Mechanism decomposition.} We compare different optimizer mechanisms in a fixed-graph, fixed-initialization framework, finding that synchrony and kernel capping alter the local--global fidelity trade-off, whereas ibFFT evaluation has small average effects on final embedding quality.
\emph{Performance and repeatability.} We evaluate end-to-end speed, fidelity, and repeatability, finding median CPU speedups of $3.29\times$ unseeded and $5.79\times$ seeded over umap-learn, with additional unseeded GPU gains over cuML on million-scale datasets. We quantify the accompanying fidelity trade-offs and fixed-seed repeatability costs. In a BRAQUE-derived workflow, we trace embedding variation to HDBSCAN assignment changes and demonstrate ibUMAP's lower CPU cost for repeatable reuse \citep{dallolio2023braque}.

\section{Background and Related Work}
\label{sec:background}

\paragraph{UMAP's optimizer is not its objective.}
UMAP belongs to the neighbor-embedding lineage that includes t-SNE \citep{vandermaaten2008visualizing}, but it builds and optimizes a sparse fuzzy neighborhood graph rather than a normalized dense input affinity matrix \citep{mcinnes2020umap}. The fuzzy cross-entropy used to motivate UMAP is not the object that the implementation optimizes: practical UMAP schedules positive edges, draws negative vertices, and applies each interaction immediately. Later events therefore observe modified coordinates, making the trajectory sensitive to event ordering.

\paragraph{Negative sampling and its expectation.}
UMAP avoids enumerating the dense non-edge term of that cross-entropy by sampling negative vertices, and analyses of this estimator motivate degree-dependent repulsive weighting \citep{damrich2021umap}. Related work in large-scale graph layout reaches a complementary result from the sampling side: SNAP-tFDP uses edge-centric negative sampling whose expected repulsive force recovers a degree-weighted full-force formulation, while retaining stochastic in-place updates \citep{chen2026snaptfdpmassivelyscalablegraph}. Changing the effective balance between attraction and repulsion moves a method along a spectrum between more continuous and more clustered representations \citep{bohm2022attraction}, and negative sampling also links UMAP to contrastive learning \citep{damrich2023contrastive}. Attraction shape further controls contraction, annealing, cluster formation, and sensitivity to initialization in UMAP-style pairwise dynamics \citep{islam2026shape}. These analyses motivate the controlled comparisons in Section~\ref{sec:field-dynamics}, which measure how optimizer mechanisms affect final embedding quality under a fixed graph and initialization.

\paragraph{Scalable \texorpdfstring{$N$}{N}-body and field evaluation.}
Dense nonlinear-embedding gradients have been expressed as $N$-body sums and accelerated with fast multipole methods \citep{vladymyrov2014linear}. GPGPU recast t-SNE repulsion as scalar and vector fields over the two-dimensional embedding domain and approximated them on GPUs using texture splatting \citep{pezzotti20gpgpu}, while FIt-SNE established the particle-to-mesh, FFT convolution, and mesh-to-particle pattern for dense neighbor-embedding forces \citep{linderman2019fast}; the Fast Kernel Transform provides a kernel-general, adaptive, accuracy-controlled alternative to a regular-grid FFT scheme \citep{ryan2022fastkernel}, and FFT/GPU field evaluation has also been applied to force-directed graph layouts \citep{zhong2023force}. Unlike t-SNE field formulations, ibUMAP targets the degree-weighted repulsive operator induced by UMAP-style optimization semantics and studies its use inside a synchronous optimizer; cuML instead accelerates the sampled UMAP pipeline on GPUs \citep{nolet2021bringing} and is our primary accelerated baseline. FastUMAP changes both the graph representation and the initialization \citep{li2026fastumap}; Appendix~\ref{app:benchmark-protocol} gives our exclusion rationale. MAP IT provides a further comparison point based on a different representation-matching objective \citep{jenssen2024mapit}.

\paragraph{What counts as a better embedding.}
Embedding fidelity has distinct local, global, density, and inter-cluster aspects, with rankings depending on preprocessing, parameters, datasets, and metrics \citep{huang2022comprehensive}. Fidelity, fixed-seed bitwise repeatability, run-to-run geometric stability, and local perturbation stability \citep{jung2026ghostumap2} address different questions. Visual artifacts and unreliable individual projections further limit interpretation \citep{bergam2026tsne,xia2024scdeed}; even diverse metrics do not establish projection accuracy \citep{machado2025necessary}.

\section{ibUMAP: Coherent and Scalable Field Evaluation}
\label{sec:method}

ibUMAP retains UMAP's graph construction and initialization algorithms and modifies only the layout-optimization stage. It keeps UMAP's attraction kernel and parameter conventions \citep{mcinnes2020umap}, replaces stochastic negative-sampling repulsion with a dense field evaluated by interpolation-based FFT (ibFFT), and applies both forces synchronously.

\subsection{Preliminaries and the degree-weighted repulsive field}
\label{sec:expected-repulsion}

Let $\{x_i\}_{i=1}^n\subset\mathbb{R}^D$ be the input and $y_i\in\mathbb{R}^2$ its embedding, and write $s_{ij}=\lVert y_i-y_j\rVert_2^2$. UMAP builds a symmetric fuzzy graph $V=(v_{ij})$ from local neighborhoods, with weighted degree $d_i^{(v)}=\sum_j v_{ij}$ and retained directed entries $\mathcal{E}$; the low-dimensional membership is $w_{ij}=(1+a s_{ij}^{b})^{-1}$, with $a,b>0$ fitted from \texttt{min\_dist} and \texttt{spread} \citep{mcinnes2020umap}. For a retained edge the optimizer applies the attractive force $f_{ij}^{a}$, and a negative pair the repulsive kernel $K_{ij}$:

\begin{equation}
    f_{ij}^{a}
    =-\frac{2ab\,s_{ij}^{b-1}}{1+a s_{ij}^{b}}(y_i-y_j),
    \qquad
    K_{ij}
    =\frac{2\gamma b}{(\varepsilon+s_{ij})(1+a s_{ij}^{b})},
    \label{eq:attractive-force}
\end{equation}

where $\gamma$ is the repulsion strength and $\varepsilon>0$ regularizes short distances \citep{mcinnes2020umap}.

For a fixed embedding $Y$, uniform negative sampling induces a deterministic conditional expectation of the sampled repulsive update. In the reference implementation this expectation depends on each head's scheduled negative-event count and on coordinate-wise clipping; Appendix~\ref{app:expectation-audit} derives it and audits the shared conventions. Motivated by degree-dependent repulsion in negative-sampling analyses \citep{damrich2021umap}, ibUMAP instead adopts the radial field

\begin{equation}
    f_{i,t}^{r,\mathrm{raw}}
    =\alpha_t E_i R_i^{\mathrm{raw}},
    \qquad
    R_i^{\mathrm{raw}}
    =\sum_{j\ne i}K_{ij}(y_i-y_j),
    \qquad
    E_i=\frac{m d_i^{(v)}}{2n},
    \label{eq:global-repulsion}
\end{equation}

where $\alpha_t$ is the learning rate, $m$ the negative-sample rate, and the factor $1/2$ offsets the head-wise attraction of Section~\ref{sec:fieldwise-optimization}, preserving the reference's expected repulsion-to-attraction ratio. This field replaces the reference implementation's time-dependent event weights and omits its per-interaction coordinate-wise clipping, so it is distinct from the exact expectation of that implementation's scheduled update; it also evaluates all interactions at one snapshot, whereas online in-place updates use successively modified coordinates. The kernel-capped field additionally limits the scalar kernel,

\begin{equation}
    K_{ij}^{\mathrm{cap}}=\min\{c_K,K_{ij}\},
    \qquad
    R_i^{\mathrm{cap}}
    =\sum_{j\ne i}K_{ij}^{\mathrm{cap}}(y_i-y_j),
    \label{eq:kernel-capped-reference-field}
\end{equation}

with $c_K=4$. The event expectation, the raw field, and the kernel-capped field are three different objects, with direct summation as the quadratic reference for the last. Kernel capping changes the field; ibFFT changes only how it is evaluated.

\subsection{Three-moment field evaluation}
\label{sec:ibfft}

Viewing a dense nonlinear-embedding gradient as an $N$-body sum follows prior work \citep{vladymyrov2014linear}. For either direct kernel $K^\star\in\{K,K^{\mathrm{cap}}\}$, the degree-weighted vector sum separates as $R_i^\star=y_iS_0(y_i)-S_1(y_i)$ with $S_0(y_i)=\sum_jK_{ij}^\star$ and $S_1(y_i)=\sum_jK_{ij}^\star y_j$. In two dimensions $S_1=(S_x,S_y)^\top$, so the field is reconstructed from three scalar moments:

\begin{equation}
    R_i^\star=
    \begin{bmatrix}
        y_{i,x}S_0(y_i)-S_x(y_i)\\
        y_{i,y}S_0(y_i)-S_y(y_i)
    \end{bmatrix}.
    \label{eq:moment-reconstruction}
\end{equation}

Every source therefore deposits the three charges $(1,y_{j,x},y_{j,y})$, which interact through the same translation-invariant scalar kernel. Self-interaction needs no explicit correction: point $i$ enters $S_0$ and $S_1$ with the same kernel weight, so its own term cancels exactly in \eqref{eq:moment-reconstruction}.

Following the interpolation-based P2M--FFT--M2P pattern established for FIt-SNE \citep{linderman2019fast}, ibFFT covers the current embedding by an adaptive square mesh with $B$ boxes per dimension and $p$ interpolation nodes per box and dimension. The mesh is rebuilt each epoch from the coordinate range, and its box width is rounded upward onto a logarithmic scale so that nearby epochs share a quantized width and can reuse a cached kernel spectrum; Appendix~\ref{app:benchmark-protocol} states the box-count rule and default grid settings. Let $\mathcal{G}(j)$ be the $p^2$ nodes associated with point $j$ and $\lambda_{jg}$ their tensor-product Lagrange weights. Particle-to-mesh deposition constructs, for $\ell\in\{0,x,y\}$,

\[
    W_\ell(g)=\sum_{j:g\in\mathcal{G}(j)}\lambda_{jg}\phi_{j,\ell},
    \quad \phi_{j,0}=1,\quad \phi_{j,x}=y_{j,x},\quad \phi_{j,y}=y_{j,y}.
\]

The chosen scalar kernel is sampled at mesh displacements and embedded into a block-circulant array of side length $L=2pB$ by mirroring and zero-padding, which yields a linear convolution without wraparound in the active region. With $M=L^2$ padded nodes and $\mathcal{K}$ the padded sampled kernel, the three mesh fields are $V_\ell=\mathcal{F}^{-1}[\mathcal{F}(\mathcal{K})\odot\mathcal{F}(W_\ell)]$. Mesh-to-particle interpolation then gives $\widetilde S_\ell(y_i)=\sum_{g\in\mathcal{G}(i)}\lambda_{ig}V_\ell(g)$, after which \eqref{eq:moment-reconstruction} reconstructs $\widetilde R_i$.

All reported configurations use $p=1$. Deposition is then an $O(n)$ accumulation into the box containing each point and mesh-to-particle interpolation is a lookup, so the scheme is piecewise constant: points sharing a box interact at the zero-displacement kernel value $2\gamma b/\varepsilon\approx1.8\times10^{3}$ (at \texttt{min\_dist=0.1}), which the cap $c_K$ bounds. The implementation supports general $p$, but raising it does not improve fidelity in our setting and costs runtime (Appendix~\ref{app:mechanism-results}), so $p>1$ is not used.

\subsection{Synchronous updates and safeguards}
\label{sec:fieldwise-optimization}

ibUMAP updates coherently: at epoch $t$, it evaluates attraction and repulsion from the same snapshot $Y^{(t)}$ and applies both only after that evaluation. Unlike in-place updates, no force sees coordinates already moved earlier in the epoch. Scheduled fuzzy-graph edges contribute \eqref{eq:attractive-force} with coordinate-wise gradient clipping; we denote the aggregated attractive contribution, including the learning-rate factor $\alpha_t$, by $f_{i,t}^{a}$, and the selected direct or ibFFT field supplies $f_{i,t}^{r}$. Production ibUMAP aggregates attraction at the head of each scheduled directed edge, whereas the controlled synchronous variants in Section~\ref{sec:field-dynamics} retain the reference two-endpoint attraction update. The aggregation convention and update order are part of the optimizer, in addition to the field operator itself.

Beyond kernel capping in \eqref{eq:kernel-capped-reference-field}, the safeguarded configuration norm-clips the repulsive update to $\tau_t=4\alpha_t$ and damps attraction at high-degree graph hubs by a factor $\eta_i\in(0,1]$ derived from the 99th percentile of weighted degree (Appendix~\ref{app:expectation-audit}). The guarded synchronous update is

\begin{equation}
    y_i^{(t+1)}=y_i^{(t)}
    +\eta_i f_{i,t}^{a}
    +\operatorname{clip}_{\lVert\cdot\rVert_2\leq\tau_t}
      \!\left(f_{i,t}^{r}\right),
    \qquad
    \alpha_{t+1}=\alpha_0\left(1-\frac{t}{T}\right).
    \label{eq:synchronous-update}
\end{equation}

The safeguards matter because online UMAP clips and applies each sampled interaction at once, whereas a synchronous step applies summed forces. Without the norm clip, degree-weighted repulsion ejects hubs at the first, attraction-free update; without damping, summed attraction throws hubs out of the main body (Appendix~\ref{app:safeguard-cases}). Section~\ref{sec:field-dynamics} measures the associated final-quality change.

\subsection{Execution paths, complexity, and repeatability}
\label{sec:method-complexity}

The CPU path uses reusable FFT plans and buffers, cached kernel spectra, and a fixed-order accumulation mode for repeatable execution. The GPU path batches the three convolution channels, reuses plans and spectra, and fuses reconstruction and update kernels where possible; FFT/GPU force-field evaluation is related to prior graph-layout systems \citep{zhong2023force}, while cuML provides the primary accelerated UMAP baseline \citep{nolet2021bringing}.

Per epoch, P2M and M2P cost $O(np^2)$, the batched convolutions cost $O(M\log M)$, and scanning the retained fuzzy graph costs $O(|\mathcal{E}|)$, giving optimizer cost $O(T[|\mathcal{E}|+np^2+M\log M])$ in time and $O(n+|\mathcal{E}|+np^2+M)$ in memory. For $p=1$, a sparse $k$-NN graph, and the nominal $M=\Theta(n)$ mesh regime, this is $O(T[nk+n\log n])$, versus $O(Tn^2)$ for the direct all-pairs field. This covers only the optimizer: graph construction and initialization are separate pipeline costs. Sampled UMAP has a different, event-count complexity, so this bound does not establish an asymptotic advantage over negative sampling; relative end-to-end performance is assessed empirically in Section~\ref{sec:scaling}.

Because the field formulation removes random negative draws, scheduled attraction, fixed-order P2M accumulation, and repeatable accelerator reductions give repeatable execution without stochastic pair selection. We define same-device repeatability under a fixed backend, precision, and reduction mode; cross-device numerical agreement depends additionally on hardware, FFT libraries, compilers, and numeric types. Section~\ref{sec:scaling} evaluates end-to-end performance, the cost of the seeded execution profile, embedding fidelity, and run-to-run stability.

\section{How Each Mechanism Changes Embedding Quality}
\label{sec:field-dynamics}

We trace how the changes from online sampled UMAP to field-based optimization affect final embedding quality, comparing eight variants on 59 datasets in the fixed-input collection, spanning tabular, image, and single-cell data with 150--94,415 observations. All variants reuse the same fuzzy graph and spectral initialization within each dataset and run for 200 epochs with 15 neighbors and \texttt{min\_dist=0.1}. We average scores over three optimizer seeds within each dataset, then weight datasets equally. Appendix~\ref{app:mechanism-results} gives the protocol and exclusion reason, variant means (Table~\ref{tab:mechanism-quality}), paired-effect summaries, and a check that weights the data families equally. We measure local fidelity by trustworthiness (TW), continuity (C), and neighborhood preservation (NP) at $k=15$ \citep{venna2006local,lee2009quality,huang2022comprehensive}, and global fidelity by random triplet accuracy (RTA) and input-to-embedding distance Spearman correlation ($\rho_D$) \citep{huang2022comprehensive}; Appendix~\ref{app:benchmark-quality} defines them and the shared sampling protocol.

The controlled sequence makes updates synchronous (A$\to$B), then replaces random negative draws with their frozen-state expectation under uniform sampling, retaining scheduled event counts and component-wise clipping (B$\to$C). D instead uses the degree-weighted radial field in \eqref{eq:global-repulsion}, changing both the weighting and the clipping convention. E--G successively add the scalar kernel cap, default FFT evaluation, and repulsion-norm clipping with attraction damping. B--G keep two-endpoint attraction, under which $E_i$ halves the reference repulsion-to-attraction ratio; H restores it with head-wise attraction and a different execution path, so G$\to$H is not single-factor (Appendix~\ref{app:expectation-audit}).

\begin{figure}[t]
  \centering
  \includegraphics[width=\linewidth]{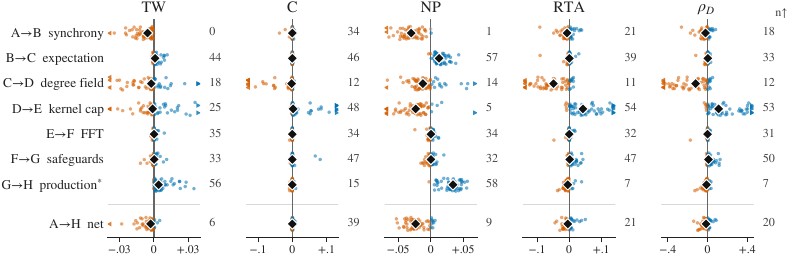}
  \caption{Paired per-dataset quality changes (later minus earlier variant; 59 datasets, seed-averaged; higher is better). Blue improves, orange worsens; diamonds are medians; numbers at right count improving datasets; triangles mark points beyond each 97.5th-percentile axis range. $^{*}$G$\to$H changes attraction update and execution path; A$\to$H is the net change from UMAP. Means: Table~\ref{tab:mechanism-quality}.}
  \label{fig:mechanism-effects}
\end{figure}

\paragraph{Synchrony and repulsion formulation.}
Synchrony lowers TW on all 59 datasets and NP on 57 (A$\to$B; Figure~\ref{fig:mechanism-effects}). The event expectation recovers about 47\% of the mean NP loss (B$\to$C). The raw degree-weighted field lowers both global scores on most datasets (C$\to$D); the kernel cap reverses this on 54 and 53 datasets and raises mean C from $0.9470$ to $0.9813$, but lowers NP on 54 (D$\to$E), trading neighbor overlap for other local and global fidelity measures.

\paragraph{FFT evaluation and update safeguards.}
FFT evaluation leaves all five paired distributions centered at zero, changing each mean by at most $0.0005$ in the local scores and $0.0016$ in $\rho_D$ (E$\to$F). Its near-field error arises from pairs sharing a mesh box, which interact at the zero-displacement kernel value and are therefore bounded by the cap $c_K$ introduced in E (Section~\ref{sec:ibfft}). Repulsion-norm clipping and attraction damping improve C, RTA, and $\rho_D$ on most datasets, raising mean $\rho_D$ by $0.0228$ at a mean NP change of $-0.0002$ (F$\to$G); Appendix~\ref{app:safeguard-cases} shows the failures they prevent.

\paragraph{Relation to the complete optimizer.}
Production ibUMAP improves NP over G on 58/59 datasets but lowers both global scores on 52/59 (G$\to$H). Relative to UMAP, it matches C but lowers mean TW, NP, RTA, and $\rho_D$ by $0.0051$, $0.0209$, $0.0037$, and $0.0179$ (A$\to$H): close to UMAP, but not equivalent. Section~\ref{sec:scaling} shows what this cost buys: faster fits and repeatable parallel CPU execution.
\FloatBarrier

\section{End-to-End Performance, Fidelity, and Repeatability}
\label{sec:scaling}

We assess the practical trade-off between end-to-end speed, embedding fidelity, and repeatability when ibUMAP is used in a complete pipeline.

\subsection{Benchmark setup}
\label{sec:experimental-protocol}

We benchmark 71 datasets with 150 to 4.17 million observations, spanning tabular data, images, word vectors, and single-cell measurements. We compare ibUMAP CPU with umap-learn \citep{mcinnes2018umap}, and ibUMAP CUDA with cuML UMAP \citep{nolet2021bringing} and TorchDR UMAP \citep{vanassel2026torchdr}. Runtime and stability comparisons retain all available pairs: 66/71/66 datasets, respectively, with five large datasets resource-skipped for CPU methods and TorchDR. The main fidelity table uses the common 66 datasets; the appendix reports all pairs.

All methods produce two-dimensional embeddings with 15 neighbors, \texttt{min\_dist=0.1}, and a requested 200-epoch budget under two profiles, \emph{unseeded} and \emph{seeded} (seed 42; deterministic ibUMAP execution). We time five runs per dataset, method, and profile after untimed warmup, including graph construction, initialization, optimization, and synchronized host materialization. Speed ratios use five-run median times and cover both profiles; main fidelity and stability comparisons are unseeded. Fidelity metrics share one sample of at most $10^6$ points (seed 42) across methods and runs. Appendix~\ref{app:benchmark-protocol} details timing, execution policies, hardware, and exclusions.

\subsection{End-to-end performance}
\label{sec:e2e-performance}

\begin{figure}[t]
  \centering
  \includegraphics[width=\linewidth]{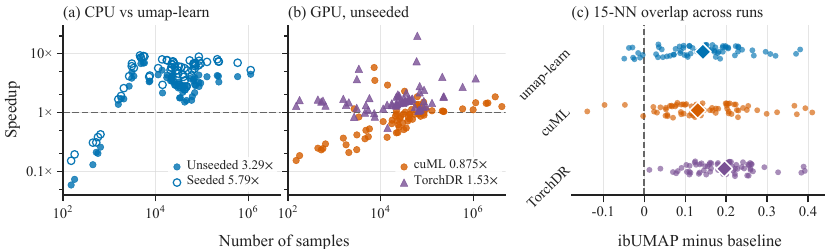}
  \caption{End-to-end speed and run-to-run stability. \textbf{(a)} CPU speedup over
  umap-learn under matching unseeded and seeded profiles and \textbf{(b)} unseeded GPU
  speedup over cuML and TorchDR: points are baseline/ibUMAP ratios of five-run median
  times and the legends give cross-dataset medians. \textbf{(c)} Paired 15-NN overlap
  differences across unseeded reruns on the same 66/71/66 pairs: positive values favor
  ibUMAP and diamonds mark cross-dataset medians. Colors denote the same baseline in all
  panels; only vertical positions in (c) are jittered.}
  \label{fig:e2e-results}
\end{figure}

\paragraph{CPU performance.}
Median speedups over umap-learn are $3.29\times$ unseeded and $5.79\times$ seeded (Figure~\ref{fig:e2e-results}a). Both profiles win on the same 60/66 datasets and lose only below 1,000 observations.

\paragraph{GPU performance.}
Unseeded ibUMAP trails cuML at the suite median (speed ratio $0.875$) but wins on all 11 datasets above $10^5$ observations, reaching a median $1.44\times$ speedup across six million-scale datasets. Against TorchDR, it wins on 62/66 datasets (median $1.53\times$ speedup; Figure~\ref{fig:e2e-results}b).

\paragraph{Cost of seeded execution.}
Both CPU implementations yield bitwise-identical embeddings across five seeded runs on all 66 datasets. Median seeded/unseeded time ratios are $0.961$ for ibUMAP and $1.664$ for umap-learn, including profile-specific threading policies. On the six million-scale GPU datasets, seeded execution substantially increases both ibUMAP and cuML runtimes, leaving ibUMAP only a $1.05\times$ median speedup. Shared cuML graph construction dominates this cost, limiting the effect of optimizer differences (Appendix~\ref{app:benchmark-runtime}, Figure~\ref{fig:appendix-stage-share}).

\FloatBarrier
\subsection{Embedding fidelity}
\label{sec:embedding-fidelity}

We use the same five fidelity measures of Section~\ref{sec:field-dynamics}, with $10^6$ sampled triplets and at most $10^6$ point pairs. Table~\ref{tab:e2e-fidelity-common} compares all methods on the common datasets under unseeded execution.

\begin{table}[htbp]
  \centering
  \small\setlength{\tabcolsep}{4pt}
  \caption{Unseeded local and global fidelity on the common 66 datasets: medians of dataset-level five-run mean scores. Metrics match Section~\ref{sec:field-dynamics}; higher is better. Bold marks the highest displayed score per CPU/GPU block, including rounding ties, not statistical significance.}
  \label{tab:e2e-fidelity-common}
  \begin{tabular}{@{}llrrrrr@{}}
    \toprule
    Platform & Method & TW $\uparrow$ & C $\uparrow$ & NP $\uparrow$ & RTA $\uparrow$ & $\rho_D$ $\uparrow$ \\
    \midrule
    CPU & umap-learn & \textbf{0.9901} & 0.9942 & \textbf{0.3699} & \textbf{0.7249} & \textbf{0.6072} \\
 & ibUMAP & 0.9853 & \textbf{0.9950} & 0.3436 & 0.7023 & 0.5493 \\
\midrule
GPU & cuML & \textbf{0.9892} & \textbf{0.9950} & 0.3635 & 0.7349 & 0.6295 \\
 & TorchDR & 0.9873 & 0.9943 & \textbf{0.3768} & \textbf{0.7647} & \textbf{0.6706} \\
 & ibUMAP & 0.9851 & \textbf{0.9950} & 0.3407 & 0.7217 & 0.5897 \\
 
    \bottomrule
  \end{tabular}
\end{table}

ibUMAP's TW, NP, RTA, and $\rho_D$ medians are lower, while C remains close, as in Section~\ref{sec:field-dynamics}. These marginal medians do not establish per-dataset advantages; paired differences, seeded scores, full coverage, and collection-weighting checks are in Appendix~\ref{app:benchmark-quality}. In return, ibUMAP fits faster on CPU, reruns more stably (Section~\ref{sec:repeatability-results}), and keeps its speed when seeded, unlike umap-learn.
The BRAQUE-derived case (Section~\ref{sec:embedding-reuse}) illustrates the lower CPU cost of obtaining fixed-seed repeatable embeddings and clustering. However, higher unseeded geometric stability does not translate into more consistent clustering in this case. Whether the runtime benefit justifies the fidelity trade-off depends on downstream requirements.

\subsection{Run-to-run embedding stability}
\label{sec:repeatability-results}

Unseeded outputs are not bitwise identical. We measure their geometric stability through run-to-run 15-nearest-neighbor overlap and sampled pairwise-distance Spearman correlation, using at most 10,000 points per comparison (Appendix~\ref{app:benchmark-stability}).

ibUMAP improves neighbor overlap on 58/66 datasets versus umap-learn, 67/71 versus cuML, and 66/66 versus TorchDR (Figure~\ref{fig:e2e-results}c). Distance correlation improves on most datasets. These gains concern geometric agreement; they imply neither input-space fidelity nor repeatable downstream partitions. Section~\ref{sec:embedding-reuse} traces how residual variation propagates to clustering.

\FloatBarrier

\section{Reusing Repeatable Embeddings}
\label{sec:embedding-reuse}

An embedding can be an input to further computation as well as a visualization. In this setting, geometric similarity across reruns may be insufficient: downstream clustering can change even when the overall layout remains similar. We examine this effect in a UMAP--HDBSCAN workflow and measure the CPU cost of making its outputs repeatable.

\subsection{Downstream clustering in a BRAQUE-derived workflow}
\label{sec:braque-case}
\label{sec:reuse-protocol}

BRAQUE uses Lognormal Shrinkage preprocessing, UMAP, and HDBSCAN clustering to aid expert interpretation of single-cell protein measurements \citep{dallolio2023braque,mcinnes2017hdbscan}. We use its public L2 sample with 56,962 cells and 62 selected features. Feature selection uses the public antibody table as a proxy for the paper-specific reference; this is a BRAQUE-derived computational case, not an exact reproduction of its biological analysis. We freeze the preprocessed input and rerun the full embedding fit in two dimensions with 50 neighbors, \texttt{min\_dist=0}, and 200 epochs. HDBSCAN settings stay fixed. Each of four CPU configurations---UMAP and ibUMAP, with and without a fixed seed---has five runs. Appendix~\ref{app:downstream-results} gives the protocol and full summary.

\begin{figure}[htbp]
  \centering
  \definecolor{ch6unchanged}{HTML}{CCD1D5}
  \definecolor{ch6cluster}{HTML}{CB6D26}
  \definecolor{ch6tonoise}{HTML}{79509A}
  \definecolor{ch6fromnoise}{HTML}{19728B}
  \begin{minipage}[b]{0.274\linewidth}
    \centering
    \includegraphics[width=\linewidth]{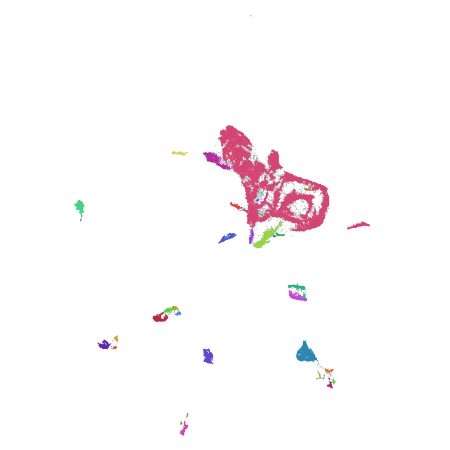}\par
    \small (a) Run 1
  \end{minipage}\hfill
  \begin{minipage}[b]{0.274\linewidth}
    \centering
    \includegraphics[width=\linewidth]{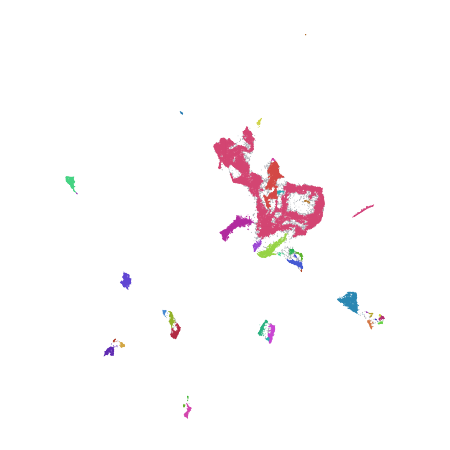}\par
    \small (b) Run 2
  \end{minipage}\hfill
  \begin{minipage}[b]{0.4247\linewidth}
    \centering
    \begin{minipage}[c]{0.645161\linewidth}
      \includegraphics[trim=0 0 118.8bp 0,clip,width=\linewidth]{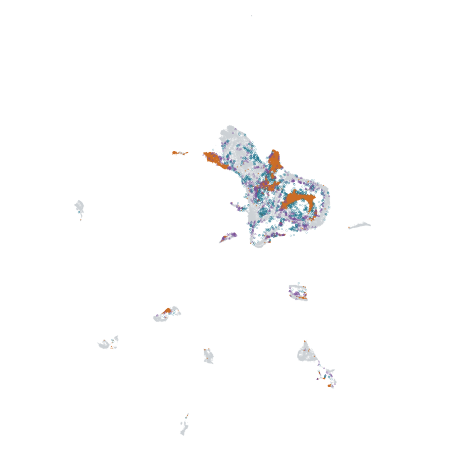}
    \end{minipage}%
    \begin{minipage}[c]{0.354839\linewidth}
      \fontsize{6.5}{9}\selectfont
      \begin{tabular}{@{}c@{\hspace{2pt}}l@{}}
        \textcolor{ch6unchanged}{$\bullet$} & Unchanged \\
        \textcolor{ch6cluster}{$\bullet$} & Cluster $\to$ cluster \\
        \textcolor{ch6tonoise}{$\triangle$} & Cluster $\to$ noise \\
        \textcolor{ch6fromnoise}{$\circ$} & Noise $\to$ cluster
      \end{tabular}
    \end{minipage}\par
    \makebox[0.645161\linewidth]{\small (c) Assignment changes}\hfill
  \end{minipage}
  \caption{The first two scheduled unseeded UMAP runs on the same frozen input, colored by HDBSCAN assignments. Cluster colors are matched by maximum cell overlap; gray denotes noise in (a,b). Run 2 is rigidly aligned to Run 1 for display, without scaling; clustering uses the original coordinates. Panel (c) locates assignment changes on Run 1, distinguishing changes between clusters from transitions to or from noise. All 56,962 cells are shown on a common coordinate scale. This pair has distance correlation $\rho=0.935$, ARI $0.615$, and 19.78\% changed assignments.}
  \label{fig:braque-repeatability}
\end{figure}

Across the ten unordered pairs of unseeded UMAP runs, sampled pairwise-distance Spearman correlation has median $0.937$, yet the median adjusted Rand index (ARI) is $0.611$ and the median changed-assignment fraction is 20.80\% after maximum-overlap, one-to-one label matching. Noise is kept separate from ordinary clusters. Figure~\ref{fig:braque-repeatability} shows the first two scheduled runs, selected without inspecting their discrepancy: 19.78\% of assignments change, including 3.49\% of cells switching noise status. Cluster counts across the five runs range from 37 to 45. These changes include splits and merges; they measure computational partition variation, not biological misclassification. Even ibUMAP's higher unseeded distance correlation ($0.999$) coexists with 22.01\% changed assignments, underscoring that geometric agreement alone does not ensure repeatable clustering.

\FloatBarrier
\subsection{Cost of repeatable reuse}
\label{sec:reuse-cost}

With seed 42 fixed, both methods produce bitwise-identical coordinates and identical HDBSCAN labels across their five runs, but at very different cost. Parallel umap-learn updates shared coordinates in a scheduling-dependent order, so seeding it falls back to one job and serial optimization to fix that order; its median embedding fit rises from 15.44\,s to 49.06\,s, a $3.18\times$ slowdown. ibUMAP applies each epoch's updates from a single snapshot and accumulates the field in a fixed order (Section~\ref{sec:fieldwise-optimization}), so parallelism does not affect the result: its deterministic mode keeps eight jobs and takes 7.10\,s, matching its unseeded time (ratio $1.001$). Seeded ibUMAP is therefore $6.91\times$ faster than seeded UMAP, or $6.42\times$ including the HDBSCAN fit (7.74\,s versus 49.70\,s); times exclude preprocessing and warmup. The speedup comes with a fidelity trade-off (Appendix~\ref{app:downstream-results}).

Repeatability also makes visual comparisons controlled. A clustering threshold can be adjusted on a cached embedding; an embedding parameter requires a refit, and with a fixed seed and deterministic execution each setting returns the same layout and partition on every request. Differences between settings then reflect the parameter rather than rerun variation, which here already changes about a fifth of the assignments. ibUMAP lowers the cost of each such refit from 49.06\,s to 7.10\,s.
\FloatBarrier

\section{Limitations and Discussion}
\label{sec:limitations}

\paragraph{Scope of the method.}
The present ibFFT construction is specialized to two-dimensional output, whereas UMAP itself is not restricted to visualization dimensions \citep{mcinnes2020umap}. Extending the field to higher dimension increases mesh and convolution costs and requires a separate accuracy and scaling study. With fixed FFT and grid-management overhead, sampled UMAP can remain preferable below the empirical crossover or when a specific fidelity metric dominates throughput.

\paragraph{Scope of the repeatability claim.}
Same-device repeatability is narrower than portable reproducibility: bitwise agreement can depend on the backend, FFT library, compiler, numeric type, and reduction order, and it does not establish GhostUMAP2-style local stability under small initialization and negative-sampling perturbations \citep{jung2026ghostumap2}. Performance and fidelity conclusions are likewise specific to the tested hardware and dataset regimes.

\paragraph{Scope of the evidence.}
Synchronous aggregation changes optimizer semantics as well as computation. Section~\ref{sec:field-dynamics} measures its final-quality effects under one fixed graph and spectral initialization, leaving robustness to other initializations untested, and whether the dynamical predictions of attraction-shape theory for pairwise UMAP \citep{islam2026shape} transfer to synchronous updates remains open. Its 59 datasets are also not independent: several single-cell subsets share a source resource, dataset size and source family are associated, and the family-weighting check limits but does not remove this dependence. We characterize the ibFFT approximation by its effect on final embeddings and by an interpolation-order sweep, rather than by field-level error against the direct sum. The downstream study uses a single BRAQUE-derived proxy workflow without expert phenotype labels; it demonstrates computational partition repeatability and its cost, while generalization across domains and effects on human interpretation remain untested.

\paragraph{Interpreting the output, and what comes next.}
One interpretive limit bears directly on ibUMAP's own output: because kernel capping and synchrony change how tightly clusters contract, area and compactness in these layouts should not be read as input density without a corresponding diagnostic \citep{narayan2021density}. The broader opportunity is to ask when other sampled interaction kernels admit a useful expected-field realization. Each such extension requires its own estimator audit, field decomposition, numerical reference, and optimizer study.

\section{Conclusion}
\label{sec:conclusion}

ibUMAP makes UMAP optimization coherent: it evaluates attraction and a degree-weighted repulsive field from one snapshot and applies them synchronously. Three scalar moments and interpolation-based FFT make this dense field cheap to evaluate on CPUs and GPUs. 
In fixed-graph, fixed-initialization comparisons on 59 datasets, this synchrony costs local fidelity, kernel capping trades neighbor overlap for continuity and global fidelity, and FFT evaluation changes average quality little. 
Coherence allows seeded CPU runs to stay parallel yet bitwise repeatable on the same device, at no median runtime penalty. 
End-to-end benchmarks yield median CPU speedups of $3.29\times$ unseeded and $5.79\times$ seeded, and a $1.44\times$ unseeded GPU speedup over cuML on million-scale datasets. In unseeded runs, ibUMAP has lower TW, NP, and global-fidelity medians but greater run-to-run stability. The BRAQUE-derived case shows that geometric stability does not ensure repeatable clustering, while ibUMAP reduces the CPU cost of repeatable fits.

\section*{AI Use Statement}

We used generative AI tools (conversational assistants, coding assistants, and agentic coding tools) for the following tasks with required disclosure.
\emph{Theoretical formulation.} We derived the expected repulsive field, its three-moment evaluation, and the UMAP kernel for interpolation-based FFT from prior work \citep{mcinnes2020umap,linderman2019fast,zhong2023force}; AI tools unified the notation and wording across our drafts of these formulations (Section~\ref{sec:method}).
\emph{Derivations.} AI tools derived the expected event counts and repulsion-to-attraction ratios in Appendix~\ref{app:expectation-audit} and implemented the accompanying audit of the reference sampling schedule; we checked the derivations and the audit results.
\emph{Hypotheses and experimental design.} We hypothesized that quality differences between ibUMAP and UMAP arise from synchronous updates, FFT approximation, and safeguards, and designed the asynchronous control, the fixed-graph and fixed-initialization protocol, and the end-to-end benchmark. AI tools helped refine this hypothesis into separate factors, including the change of field formulation, and proposed variants C and D of the ablation sequence (Section~\ref{sec:field-dynamics}).
\emph{Implementation and data preparation.} We wrote the UMAP kernel and the overall architecture ourselves, adapting the FFT kernels of \citet{zhong2023force} and the umap-learn pipeline. AI tools migrated and consolidated this code into the released repository and unified our earlier dataset download and preprocessing scripts. The live demo of the downstream case was implemented by AI (Appendix~\ref{app:live-demo}).
We did not use generative AI tools to generate synthetic datasets, to formulate the main claims (replacing negative sampling with an expected field and evaluating it by FFT), or to select evaluation metrics. Translation and qualitative data analysis are not applicable to this work.
Additionally, we used AI tools to search for and summarize related work, including identifying the BRAQUE case, to brainstorm and organize ideas, to edit code, to create and revise figures, and to draft and polish parts of the paper.
In the accompanying video (linked from the repository), the narration was synthesized by an AI text-to-speech tool from a script that AI tools helped edit; the slides, animations, and demo screen recordings were made by us without generative AI.
The interpretations of results are ours; AI tools drafted part of the interpretive text, which we reviewed and revised.
The authors reviewed all AI-assisted work and checked every cited reference against the original publication. All reported numbers, including runtimes and quality scores, are produced by experiment scripts from recorded run outputs.
We verified the migrated code with the repository's unit and protocol tests, by checking that it rebuilds the fixed inputs of the mechanism study bitwise and rescores a sample of stored embeddings to the reported values, and by regenerating all result tables and figures from the frozen run summaries.
We take responsibility for the final content of this work, including text, claims, and artifacts produced with the aid of generative AI.

\section*{Ethics Statement}

This work involves no human subjects and no new data collection. All datasets are previously published and publicly available; human-derived single-cell data are used only in the form released by their providers. Our repository provides scripts that download the data from their original sources rather than redistributing them.
Faster or bitwise-repeatable embeddings are not thereby more faithful: ibUMAP retains fidelity trade-offs (Section~\ref{sec:embedding-fidelity}), and two-dimensional embeddings can mislead interpretation (Section~\ref{sec:limitations}).

\section*{Reproducibility Statement}

Section~\ref{sec:method} specifies ibUMAP, and Section~\ref{sec:method-complexity} its execution paths and repeatability modes. Appendix~\ref{app:expectation-audit} derives the expected field and audits it against the reference implementation. Appendices~\ref{app:mechanism-results}, \ref{app:benchmark-results}, and \ref{app:downstream-results} give the protocols of the mechanism study, the frozen end-to-end benchmark (including software versions, hardware, and the complete dataset manifest), and the downstream case. The repository \url{https://github.com/Bachery/ibUMAP} contains the ibUMAP source code, environment specifications, data download and preprocessing scripts, experiment and plotting scripts, and the result tables from which the reported numbers and figures are generated, as well as a live demo of the downstream case (Appendix~\ref{app:live-demo}).

\bibliography{references}
\bibliographystyle{iclr2027_conference}

\clearpage
\appendix

\section{Expected Field: Derivation and Implementation Audit}
\label{app:expectation-audit}

This appendix derives the frozen-state expectation of UMAP's sampled repulsive update, relates the degree-weighted field of \eqref{eq:global-repulsion} to it, and audits the conventions that the variants of Section~\ref{sec:field-dynamics} share or change. Statements about the reference refer to \texttt{umap-learn} 0.5.12 (its sampling calendar and single-epoch Euclidean optimizer); statements about ibUMAP refer to its CPU and CUDA optimizers and to the code of the controlled variants. The calendar statistics in Table~\ref{tab:calendar-audit} replay the reference sampling schedule, without random draws, on the saved optimizer graphs of the 59 mechanism datasets; they involve neither optimization nor quality evaluation.

\paragraph{Graph and edge counting.}
The optimizer receives the symmetric fuzzy graph after UMAP's pruning step, which removes entries with $v_{ij}<v_{\max}/T$. Each undirected edge is stored as two directed entries $(i,j),(j,i)\in\mathcal{E}$ of equal weight, and ibUMAP and the controlled variants both take $d_i^{(v)}$ as the row sum of the pruned graph. All 59 saved graphs are exactly symmetric, have $v_{\max}=1$ and $v_{\min}/v_{\max}\geq 1/T=0.005$, and contain no isolated vertex. An entry has the sampling interval $\tau_{ij}=v_{\max}/v_{ij}\geq1$: it is active at epoch $t$ once its next-sample time is at most $t$, after which that time advances by $\tau_{ij}$. Its long-run activation rate is therefore $v_{ij}/v_{\max}$ per epoch, and head $i$ has on average $d_i^{(v)}/v_{\max}$ active entries per epoch.

\paragraph{Activation schedule.}
Next-sample times start at $\tau_{ij}\geq1$, so no entry is active at epoch $0$; the first active epoch is $t=1$ on all 59 datasets. Each entry also carries a negative-sample clock $\sigma_{ij}$ with interval $\tau_{ij}/m$, initialized to $\tau_{ij}/m$. When entry $(i,j)$ is active at epoch $t$, it draws $\lfloor m(t-\sigma_{ij})/\tau_{ij}\rfloor$ negative samples, and $\sigma_{ij}$ advances by that many intervals. A first activation thus draws between $m-1$ and $2m-2$ negatives, and later activations draw $m$ on average. Summing over the entries with head $i$ gives the scheduled count $c_i(t)$. Its time average $\bar c_i$ is close to $m d_i^{(v)}/v_{\max}$, slightly lower because epoch $0$ is inactive and first activations draw fewer negatives (Table~\ref{tab:calendar-audit}). The total count also varies over time: at $t=1$ only entries with $v_{ij}=v_{\max}$ are active and the total is about $0.17$ of its mean over $t\geq1$, whereas from $t=10$ onward it stays within $0.91$--$1.10$ of that mean on every dataset.

\paragraph{Negative samples and the event expectation.}
A negative sample for head $i$ is drawn as $k=\texttt{tau\_rand\_int}\bmod n$ from a per-head Tausworthe state. We treat $k$ as uniform on all $n$ points, including $i$, and ignore the modulo bias of the integer generator. A self-draw is skipped but still counts toward $c_i(t)$; a draw at a coincident non-self point contributes zero, because its coefficient is set to zero. Only the head moves under repulsion. The online reference applies each draw to the head's current, already modified coordinates, whereas in the synchronous variant B all draws of an epoch read the snapshot $Y^{(t)}$. Conditional on $Y^{(t)}$ and $c_i(t)$, the expected repulsive update of head $i$ in B is
\begin{equation}
  f_{i,t}^{r,C}=\frac{\alpha_t c_i(t)}{n}
    \sum_{j\ne i}\operatorname{clip}_{[-4,4]}
       \bigl(K_{ij}(y_i-y_j)\bigr),
  \label{eq:scheduled-repulsion}
\end{equation}
where clipping acts on each coordinate of each interaction and $K_{ij}$ is the kernel of \eqref{eq:attractive-force} with $\gamma=1$ and the reference's hard-coded $\varepsilon=10^{-3}$. Variant C evaluates \eqref{eq:scheduled-repulsion} exactly, with \texttt{float64} accumulation, and reads $c_i(t)$ from the shared sampling calendar before the calendar advances. A unit test enumerates every possible negative target of a single sampling event and reproduces \eqref{eq:scheduled-repulsion} to within $10^{-12}$.

\paragraph{Update symmetry and attraction.}
In the reference, an active entry $(i,j)$ moves both endpoints: $y_i$ by $\alpha_t\operatorname{clip}_{[-4,4]}(f_{ij}^{a})$ and $y_j$ by the negative of that amount. Because $(i,j)$ and $(j,i)$ are both stored and follow the same schedule, each point receives on average $2d_i^{(v)}/v_{\max}$ attractive events per epoch, half as head and half as tail; the calendar replay confirms that head and tail event counts coincide for every point. Variants B--G reuse this two-endpoint update. Production ibUMAP (H) accumulates each active entry only at its head, and the mirrored entry supplies the other endpoint, so each point receives $d_i^{(v)}/v_{\max}$ attractive events per epoch, half the reference count. The attractive coefficient, its coordinate-wise clip at $4$, and the parameters $a$ and $b$, fitted by the reference routine from \texttt{min\_dist} and \texttt{spread}, are identical in all variants.

\paragraph{Normalization of the degree-weighted field.}
D--H replace the event weight $\alpha_t c_i(t)/n$ by $\alpha_t E_i$ with $E_i=m d_i^{(v)}/(2n)$. This weight is constant over epochs and, since $v_{\max}=1$, about half of the average event weight: $E_i/(\bar c_i/n)$ has median $0.507$ (Table~\ref{tab:calendar-audit}). In expectation, the ratio of repulsive weight to attractive events per point is $(m d_i^{(v)}/n)/(2d_i^{(v)})=m/(2n)$ for the reference, B, and C; $(m d_i^{(v)}/(2n))/(2d_i^{(v)})=m/(4n)$ for D--G; and $(m d_i^{(v)}/(2n))/d_i^{(v)}=m/(2n)$ for H.
Combined with head-wise attraction, the factor $1/2$ therefore preserves the reference balance between repulsion and attraction and halves both, which in expectation amounts to halving the step size. Under the two-endpoint attraction of D--G, the same factor halves the relative repulsion. Measured on the replayed calendars, this balance relative to the reference has median $0.507$ for D--G and $1.014$ for H. Consequently, C$\to$D lowers the average repulsive weight by about half, in addition to removing its time dependence and the component-wise clip; D--G run at about half the reference balance; and G$\to$H restores it while also changing the execution path. These are statements about expected event weights. Clipping, the kernel cap, and damping further change the forces, and no variant isolates the factor experimentally.

\begin{table}[H]
  \centering\small
  \caption{Replayed sampling calendars on the 59 mechanism datasets ($T=200$, $m=5$). Per-point quantities are summarized by their median within each dataset; epoch totals are per dataset. The table gives the median and range of these values across datasets.}
  \label{tab:calendar-audit}
  \begin{tabular}{@{}lrr@{}}
    \toprule
    Quantity & Median & Range \\
    \midrule
    Mean scheduled count, $\bar c_i/(m d_i^{(v)})$ & $0.986$ & $0.984$--$0.987$ \\
First activations drawing $m-1$ negatives (share of entries) & $0.766$ & $0.728$--$0.862$ \\
Total count at $t=1$ / mean over $t\geq1$ & $0.173$ & $0.163$--$0.261$ \\
Minimum total count over $t\geq10$ / mean over $t\geq1$ & $0.983$ & $0.907$--$0.989$ \\
Maximum total count over $t\geq1$ / mean over $t\geq1$ & $1.016$ & $1.012$--$1.099$ \\
$E_i/(\bar c_i/n)$, equal to the D--G balance / reference & $0.507$ & $0.507$--$0.508$ \\
Repulsion-to-attraction balance of H / reference & $1.014$ & $1.013$--$1.017$ \\
 
    \bottomrule
  \end{tabular}
\end{table}

\paragraph{Kernel parameters and clipping.}
A and B clip each coordinate of each sampled repulsive interaction to $[-4,4]$, and C applies the same clip inside its expectation. D applies no repulsive clip. E--H cap the scalar kernel at $c_K=4$ and do not clip individual interactions. The two conventions are not nested: the component clip bounds each interaction's contribution $K_{ij}\lvert y_{i,d}-y_{j,d}\rvert$, whereas the cap bounds $K_{ij}$ and hence limits the contribution to $4\lVert y_i-y_j\rVert_2$, which is stricter for nearby points. G and H additionally norm-clip the aggregated repulsive update at $4\alpha_t$ and damp attraction by $\eta_i=\min\{1,(d_{99}/d_i^{(v)})^{1/2}\}$, where $d_{99}$ is the 99th percentile of weighted degree.

\paragraph{Learning-rate schedule and the first update.}
All variants update the learning rate after each epoch, $\alpha_{t+1}=\alpha_0(1-t/T)$, so epochs $0$ and $1$ both use $\alpha_0$ and epoch $T-1$ uses $2\alpha_0/T$. Because epoch $0$ is inactive in the reference and in A--C, their first update occurs at $t=1$; in C the recorded scheduled count at $t=0$ is zero on all datasets. D--H evaluate the dense field at every epoch, including $t=0$, when no attractive entry is active yet: their first update is purely repulsive and uses the full initial learning rate. Production ibUMAP evaluates its field once per epoch, $T$ times in total. Most recorded activity of the repulsion-norm clip in G falls in this update: it clips a median of $26.6\%$ of points (range $0\%$--$95.1\%$ across the 59 datasets), whereas at the later recorded steps, taken every five epochs, the largest clipped share has median $0\%$ and never exceeds $9.7\%$.

\paragraph{Precision and implementation checks.}
Embeddings are \texttt{float32} in all variants. CPU sampling calendars use \texttt{float64}, as in the reference, whereas CUDA calendars use \texttt{float32}, which can shift individual activation epochs. Direct field sums in C--E accumulate in \texttt{float64}. The asynchronous implementation control reproduces reference UMAP bitwise in all 177 retained dataset--seed comparisons, and the synchronous routine shared by B--G is checked bitwise against the library implementation. The derivation concerns a frozen state and ideal uniform sampling; it does not imply that the trajectories of different variants agree.
\FloatBarrier

\section{Mechanism Study: Protocol and Robustness}
\label{app:mechanism-results}

\paragraph{Coverage and weighting.}
The fixed-input collection contains 60 candidate datasets with at most $10^5$ observations. We retain 59: 11 tabular datasets, 4 image datasets, 30 Tabula Sapiens subsets, 4 scDEED datasets, and 10 C.~elegans subsets. The retained collection spans 150--94,415 observations and includes Spambase and HIVA. Dataset identifiers and scores are retained in the machine-readable experiment outputs.

We exclude the C.~elegans global-QC annotated dataset ($n=54{,}649$) from all variants and seeds because its asynchronous implementation control fails the bitwise comparison with reference UMAP at seed 2026. Two repeated checks on the original CPU reproduce each implementation's saved output but confirm the disagreement between them; its cause remains unresolved. This exclusion preceded quality evaluation. The raw outputs and diagnostic evidence are retained.

These datasets are not 59 independent data collections: several single-cell subsets share their source resource or observations, and data size is associated with source family. We therefore avoid interpreting coverage differences as isolated size effects. Table~\ref{tab:mechanism-quality} weights datasets equally, while Table~\ref{tab:mechanism-family} first averages datasets within each family and then weights the five families equally. The latter is a sensitivity check, not a claim that the families define a uniquely correct target population.

\paragraph{Matched optimization.}
All variants consume saved fuzzy graphs and spectral initial coordinates constructed with seed 42. Common parameters are two output dimensions, Euclidean distance, 15 neighbors, \texttt{min\_dist=0.1}, spread 1, learning rate 1, repulsion strength 1, negative-sample rate 5, and 200 epochs. CPU runs use \texttt{float32} embeddings, eight configured Numba/FFT threads, \texttt{umap-learn} 0.5.12, and FFTW's estimate planner. Direct field sums accumulate in \texttt{float64}. The matched variants use the reference post-epoch learning-rate schedule, so epochs 0 and 1 both use the initial learning rate; epoch 0 applies no attraction (Appendix~\ref{app:expectation-audit}). The additional asynchronous implementation control is bitwise identical to reference UMAP in all 177 retained dataset--seed comparisons and is not separately scored.

\paragraph{Variant semantics.}
C uses the ideal-uniform frozen-state expectation \eqref{eq:scheduled-repulsion}, with the scheduled negative-event count $c_i(t)$ read from the shared sampling calendar before its advancement; Appendix~\ref{app:expectation-audit} derives it and audits the shared conventions. D instead uses $\alpha_t m d_i^{(v)}/(2n)$ times the raw radial-field sum, without component-wise clipping. E applies the scalar kernel cap $c_K=4$. F uses the default piecewise-constant FFT setting ($p=1$ and box-scale multiplier 1) described in Appendix~\ref{app:benchmark-protocol}. G additionally clips the repulsive update at norm $4\alpha_t$ and damps attraction using the 99th-percentile weighted degree and exponent 0.5. Their joint comparison does not isolate the two update safeguards individually. H uses the production optimizer, including its head-wise attraction aggregation. The additional B~+~repulsion-norm clip branch retains sampled repulsion and adds only the per-epoch repulsion-norm threshold $4\alpha_t$.

\paragraph{Scores and repeated records.}
Seeds 42, 137, and 2026 vary optimizer draws only; the graph and initialization stay fixed. C--G do not depend on these draws, and their deterministic trajectories are computed once and reused. Within the retained set, the 1,770 optimization records comprise 1,180 executed runs and 590 deterministic reuse records. All five quality metrics cover the 1,593 evaluation records for A--H and the sampled-repulsion clipping control. These counts do not represent independent repetitions for the deterministic variants.

All rows of each dataset enter quality evaluation. TW, C, and NP use 15 neighbors; RTA and $\rho_D$ use the million-triplet/pair protocol with sampling seed 42 in Appendix~\ref{app:benchmark-quality}. The paired-effect and family-weighting tables report all five metrics; the paired table also includes the additional clipping control. We report descriptive comparisons without noninferiority or significance claims.

\begin{table}[H]
  \centering\small
  \setlength{\tabcolsep}{4pt}
  \caption{Local (TW, C, NP) and global (RTA, $\rho_D$) fidelity under a fixed graph and initialization: means across 59 datasets after averaging optimizer seeds within each dataset. Higher is better. H, the production optimizer used in Sections~\ref{sec:scaling} and~\ref{sec:embedding-reuse}, differs from G in both attraction aggregation and execution path; its net effect is read against A (UMAP), not as one mechanism after G.}
  \label{tab:mechanism-quality}
  \begin{tabular}{@{}lrrrrr@{}}
    \toprule
    Variant & TW $\uparrow$ & C $\uparrow$ & NP $\uparrow$ & RTA $\uparrow$ & $\rho_D$ $\uparrow$ \\
    \midrule
    A. UMAP & 0.9718 & 0.9834 & 0.4281 & 0.7070 & 0.5627 \\
B. Synchronous sampled & 0.9630 & 0.9825 & 0.3973 & 0.7006 & 0.5366 \\
C. Synchronous event expectation & 0.9648 & 0.9829 & 0.4118 & 0.7001 & 0.5349 \\
D. Direct degree-weighted field & 0.9568 & 0.9470 & 0.4010 & 0.6463 & 0.3583 \\
E. D + kernel cap & 0.9594 & 0.9813 & 0.3756 & 0.7013 & 0.5396 \\
F. E with default FFT & 0.9598 & 0.9812 & 0.3760 & 0.7012 & 0.5379 \\
G. F + norm clip and damping & 0.9599 & 0.9841 & 0.3758 & 0.7083 & 0.5608 \\
\midrule
H. Production ibUMAP & 0.9667 & 0.9838 & 0.4072 & 0.7033 & 0.5448 \\
 
    \bottomrule
  \end{tabular}
\end{table}

\begin{table}[H]
  \centering\footnotesize
  \setlength{\tabcolsep}{3pt}
  \caption{Distribution of paired quality changes across 59 datasets. Each cell gives the median signed quality improvement and, in parentheses, the number of datasets with an improvement, computed before rounding. Positive values favor the later variant. A--H match Table~\ref{tab:mechanism-quality}, and Figure~\ref{fig:mechanism-effects} plots the underlying per-dataset changes. The final row is the additional sampled-repulsion clipping control.}
  \label{tab:mechanism-effects}
  \begin{tabular}{@{}lrrrrr@{}}
    \toprule
    Comparison & TW & C & NP & RTA & $\rho_D$ \\
    \midrule
    B--A & $-0.0055$ (0) & $+0.0001$ (34) & $-0.0302$ (1) & $-0.0064$ (21) & $-0.0211$ (18) \\
C--B & $+0.0010$ (44) & $+0.0001$ (46) & $+0.0128$ (57) & $+0.0007$ (39) & $+0.0023$ (33) \\
D--C & $-0.0023$ (18) & $-0.0018$ (12) & $-0.0121$ (14) & $-0.0489$ (11) & $-0.1211$ (12) \\
E--D & $-0.0011$ (25) & $+0.0022$ (48) & $-0.0232$ (5) & $+0.0431$ (54) & $+0.1112$ (53) \\
F--E & $+0.0001$ (35) & $+0.0000$ (34) & $+0.0004$ (34) & $+0.0000$ (32) & $+0.0001$ (31) \\
G--F & $+0.0002$ (33) & $+0.0001$ (47) & $+0.0002$ (32) & $+0.0017$ (47) & $+0.0089$ (50) \\
H--G & $+0.0041$ (56) & $-0.0001$ (15) & $+0.0344$ (58) & $-0.0042$ (7) & $-0.0113$ (7) \\
\midrule
H--A & $-0.0028$ (6) & $+0.0001$ (39) & $-0.0233$ (9) & $-0.0043$ (21) & $-0.0152$ (20) \\
\midrule
B + norm clip -- B & $+0.0001$ (33) & $+0.0001$ (44) & $-0.0008$ (25) & $+0.0055$ (46) & $+0.0125$ (45) \\
 
    \bottomrule
  \end{tabular}
\end{table}

\paragraph{Complete optimizer versus UMAP.}
Because H differs from G in more than one factor, its comparison with A gives the net effect of all changes. Production ibUMAP matches UMAP in C (mean $+0.0004$, 39/59 datasets improving) and trails it in the other four metrics: mean TW, NP, RTA, and $\rho_D$ are lower by $0.0051$, $0.0209$, $0.0037$, and $0.0179$, improving on 6/59, 9/59, 21/59, and 20/59 datasets (Table~\ref{tab:mechanism-effects}). The largest gaps, in NP and $\rho_D$, amount to about 5\% and 3\% of UMAP's means. Under matched inputs, the complete optimizer is therefore close to UMAP but not equivalent; Section~\ref{sec:scaling} reports the speed and repeatability obtained in exchange.

\paragraph{Heterogeneity of effects.}
The median paired differences in Table~\ref{tab:mechanism-effects} complement the means in the main table. For example, the kernel cap increases mean trustworthiness but decreases its median paired value; it does not improve this metric uniformly. The default FFT comparison has a median absolute change of $0.0018$ in NP. Repulsion-norm clipping and attraction damping improve C on 47/59 datasets, while their NP effects are mixed. Adding norm clipping to sampled repulsion alone improves RTA on 46/59 datasets and $\rho_D$ on 45/59, with median gains of $0.0055$ and $0.0125$.

\begin{table}[t]
  \centering\small
  \setlength{\tabcolsep}{5pt}
  \caption{Safeguard failure cases on the fixed inputs of the mechanism study (production optimizer H, one safeguard switched off at a time; seed 42, as all three seeds give bitwise-identical embeddings). Ratio: bounding-box area of all points over that of the 99\% of points nearest the median. Far points lie beyond twice the 99\% radius. The peak is the largest ratio after any of the 200 updates, with the update index in parentheses; the last column counts updates after which the ratio exceeds 10.}
  \label{tab:safeguard-cases}
  \begin{tabular}{@{}llrrrr@{}}
    \toprule
    Dataset & Variant & Final ratio & Far points & Peak ratio & Updates $>10$ \\
    \midrule
    CIFAR-10 & All safeguards & 1.46 & 0 & 11.8 (7) & 12 \\
 & No repulsion-norm clip & 1{,}940 & 555 (0.9\%) & 1{,}940 (200) & 200 \\
 & No attraction damping & 1.70 & 0 & 90.2 (2) & 157 \\
\midrule
scDEED CART & All safeguards & 1.02 & 0 & 18.6 (21) & 38 \\
 & No repulsion-norm clip & 93.9 & 483 (0.8\%) & 93.9 (200) & 146 \\
 & No attraction damping & 2.08 & 1 ($<$0.1\%) & 296 (17) & 175 \\
 
    \bottomrule
  \end{tabular}
\end{table}

\begin{figure}[t]
  \centering
  \includegraphics[width=\linewidth]{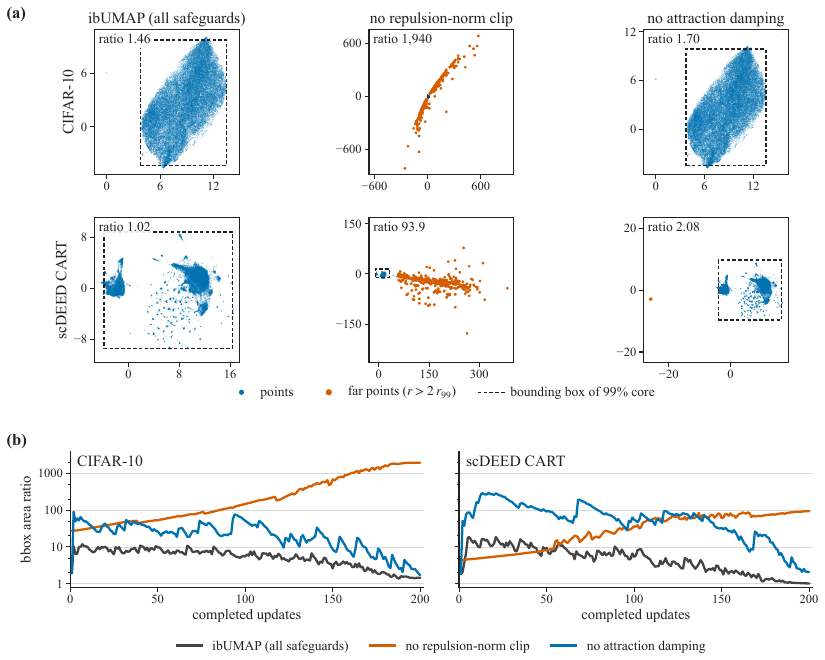}
  \caption{Safeguard failure cases. \textbf{(a)} Final embeddings of production ibUMAP and of the same optimizer without the repulsion-norm clip or without attraction damping. Each panel spans the full extent of its embedding at equal aspect; orange marks far points, the dashed rectangle bounds the 99\% core, and the inset gives the bounding-box area ratio. \textbf{(b)} The same ratio after every update. Without the clip, escaped points persist to the end; without damping, displaced hubs dominate the layout for most of the run and return only as the learning rate decays.}
  \label{fig:safeguard-cases}
\end{figure}

\paragraph{Safeguard failure cases.}
\label{app:safeguard-cases}
Repulsion-norm clipping and attraction damping were introduced after individual runs failed. Figure~\ref{fig:safeguard-cases} and Table~\ref{tab:safeguard-cases} revisit these runs with the production optimizer H, switching off one of the two safeguards at a time; the kernel cap stays on. CIFAR-10 and scDEED CART reuse the fixed graphs and initializations of this study, and all other settings match H. Under these inputs the three optimizer seeds give bitwise-identical embeddings for every variant, and the complete variant reproduces the H embeddings used above bitwise. We summarize a layout by the ratio of its bounding-box area to that of the 99\% of points nearest the coordinate-wise median, and call a point far when its distance to the median exceeds twice that of the 99\% quantile. On the 59 datasets, H gives a median ratio of 1.01 and a maximum of 1.52, UMAP a maximum of 1.93, and neither leaves a far point.

Without the repulsion-norm clip, points escape at the first update, which is purely repulsive and uses the full learning rate (Appendix~\ref{app:expectation-audit}). Its largest step is 897 on CIFAR-10 and 418 on scDEED CART, whereas the clip limits it to $4\alpha_0=4$. The escaped points are mostly graph hubs, with a median weighted-degree percentile above 99, consistent with the degree weight $E_i$ in \eqref{eq:global-repulsion}. They keep drifting outwards, so 555 of 60{,}000 and 483 of 62{,}167 points remain far in the final layouts, and the ratio reaches 1{,}940 and 93.9: the main body fills only a corner of a default scatter plot (Figure~\ref{fig:safeguard-cases}a).

Without attraction damping, the failure appears during optimization rather than in the final layout. Attraction accumulated over many edges reaches norm 455 at a scDEED CART hub, against 102 with damping, and throws hubs out of the main body: the ratio exceeds 10 after 175 of the 200 updates and peaks at 296 (Figure~\ref{fig:safeguard-cases}b). While the hubs are displaced, the adaptive mesh must also cover them, and the longer side of the 99\% core spans a median of 28.5 mesh boxes over the run, against 61.3 with damping. As the learning rate decays, the hubs are pulled back. Only the vertex of largest weighted degree (1{,}999, against a 99th percentile of 45.7) remains far, and the final ratio of 2.08 still exceeds that of every H and UMAP layout in the collection. CIFAR-10 behaves alike, with a ratio above 10 after 157 updates and a final ratio of 1.70. Damping thus keeps the layout from depending on this late recovery; smaller transient excursions remain with both safeguards, with peak ratios of 11.8 and 18.6.

These cases motivate the safeguards but do not calibrate them. The thresholds, $4\alpha_t$ for the clip and the 99th percentile with exponent 0.5 for damping, were fixed while developing an earlier version of the optimizer, and we have not repeated that selection for the final version.

\paragraph{Interpolation order.}
The default ibFFT setting uses $p=1$. A separate optimization-only benchmark
(30 datasets from six families, 150--129,062 observations, fixed fuzzy graph and
spectral initialization, boundary \texttt{optimize\_from\_graph}) compares fixed
$p=1$, $p=2$, and $p=3$ on CPU and CUDA. Table~\ref{tab:mechanism-psweep} reports
paired per-dataset differences against $p=1$. Median changes do not exceed
$4\times10^{-4}$ on any of the three local metrics and the improvement counts
stay near half the datasets, while the CPU optimizer time rises by 10.3\% at
$p=2$ and 23.9\% at $p=3$. A persistent $p1\!\to\!p2\!\to\!p3$ stage schedule
behaves likewise.
Within this 30-dataset sweep, increasing the interpolation order offers no consistent improvement in these three local metrics. Together with the small average E$\to$F differences in Table~\ref{tab:mechanism-quality}, this suggests limited final-quality benefit from higher interpolation orders in the tested configurations, but does not establish the numerical accuracy of the approximate field.
Three caveats apply. This benchmark predates the
reported implementation and its metric set differs from the main study, which
adds RTA and $\rho_D$; its three repeats produce bitwise-identical embeddings for
every deterministic variant, so each dataset contributes one quality record and
three timing replicates; and $p>1$ also forgoes the $p=1$ particle-to-mesh path
and enlarges the padded mesh to $M=(2pB)^2$, so the runtime ratios combine
interpolation arithmetic, mesh growth, and deposition mode.

\begin{table}[H]
  \centering\small
  \setlength{\tabcolsep}{5pt}
  \caption{Interpolation-order sweep on 30 datasets, paired against $p=1$. Each
  quality cell gives the median signed difference and, in parentheses, the number
  of datasets improved. The last column is the median ratio of optimizer wall time
  to $p=1$; values above one indicate a slowdown.}
  \label{tab:mechanism-psweep}
  \begin{tabular}{@{}llrrrr@{}}
    \toprule
    Device & Order & TW & C & NP & Time \\
    \midrule
    CPU & $p=2$ & $+0.00000$ (16/30) & $+0.00001$ (18/30) & $+0.00025$ (16/30) & $1.103\times$ \\
 & $p=3$ & $-0.00005$ (13/30) & $+0.00002$ (19/30) & $-0.00013$ (14/30) & $1.239\times$ \\
\midrule
CUDA & $p=2$ & $-0.00009$ (11/30) & $+0.00002$ (21/30) & $-0.00036$ (13/30) & $1.012\times$ \\
 & $p=3$ & $-0.00003$ (13/30) & $+0.00001$ (20/30) & $-0.00003$ (14/30) & $1.019\times$ \\
 
    \bottomrule
  \end{tabular}
\end{table}

\begin{table}[H]
  \centering\small
  \setlength{\tabcolsep}{6pt}
  \caption{Family-weighting sensitivity check. Entries average dataset scores within each of the five families, then average the families equally. Variant IDs and metric directions match Table~\ref{tab:mechanism-quality}; deterministic seed reuse is handled identically.}
  \label{tab:mechanism-family}
  \begin{tabular}{@{}lrrrrr@{}}
    \toprule
    Variant & TW $\uparrow$ & C $\uparrow$ & NP $\uparrow$ & RTA $\uparrow$ & $\rho_D$ $\uparrow$ \\
    \midrule
    A & 0.9541 & 0.9758 & 0.3413 & 0.6989 & 0.5420 \\
B & 0.9460 & 0.9728 & 0.3182 & 0.6941 & 0.5213 \\
C & 0.9485 & 0.9734 & 0.3327 & 0.6940 & 0.5202 \\
D & 0.9377 & 0.9149 & 0.3194 & 0.6405 & 0.3401 \\
E & 0.9416 & 0.9683 & 0.2968 & 0.6956 & 0.5269 \\
F & 0.9421 & 0.9683 & 0.2978 & 0.6960 & 0.5258 \\
G & 0.9417 & 0.9758 & 0.2968 & 0.7009 & 0.5445 \\
\midrule
H & 0.9504 & 0.9761 & 0.3274 & 0.6971 & 0.5307 \\
 
    \bottomrule
  \end{tabular}
\end{table}

Under equal family weights, synchrony's NP change remains negative ($-0.0231$), the FFT change remains small ($+0.0010$), and production ibUMAP remains below UMAP on NP ($-0.0139$) and both global scores. The kernel cap still raises mean C while lowering mean NP under family weighting. Family weighting limits dominance by numerous related subsets but does not remove all dependence, and the image family remains limited to four datasets. The fixed spectral initialization also leaves generalization to other initializations untested.
\FloatBarrier

\section{End-to-End Benchmark: Protocol and Additional Results}
\label{app:benchmark-results}

\subsection{Protocol, implementation profiles, and data coverage}
\label{app:benchmark-protocol}

All fits in Section~\ref{sec:scaling} follow the frozen end-to-end benchmark protocol \texttt{20260806.1}, published on 9 August 2026. It contains 3,400 successful runs and 10,200 valid local-fidelity records. We additionally evaluate RTA and $\rho_D$ on all 1,700 saved unseeded embeddings and the first seeded repeat of each dataset--method pair (340 embeddings), adding 4,080 global-fidelity records without refitting. All scheduled fits completed successfully. The nominal grid is 71 datasets, 5 implementations, 2 execution profiles, and 5 repetitions. The implementations are umap-learn, ibUMAP CPU, cuML UMAP, ibUMAP CUDA, and TorchDR UMAP. The 150 resource-skipped combinations account for the difference between the nominal 3,550 runs and the 3,400 executed runs.

\paragraph{Baseline scope.}
FastUMAP accelerates the upstream representation through a point--landmark fuzzy graph and a Nystr{\"o}m spectral warm start \citep{li2026fastumap}, changing both the graph representation and the initialization. We exclude it because its preprocessing, threading, landmark ratio, timing boundary, and quality evaluation cannot be matched independently of those upstream changes.

\paragraph{Coverage and exclusions.}
The task builder excludes both profiles of umap-learn, ibUMAP CPU, and TorchDR on GIST, Google News, and MERFISH animals 1, 2, and 3. These exclusions were configured because of known host or device memory exhaustion. They are \emph{resource skips}, not observed failures among the scheduled v3 runs, and do not establish a memory-capacity advantage from a controlled memory benchmark. cuML and ibUMAP CUDA both cover all 71 datasets. CPU and TorchDR comparisons cover 66, reaching 1,136,218 observations. Table~\ref{tab:benchmark-datasets} lists every dataset, including subsets derived from the same underlying collection.

Table~\ref{tab:e2e-fidelity-common} uses the intersection of datasets with completed runs for all five implementations: the 66 datasets outside these five exclusions. The same set is available under both execution profiles. This common set supports a five-method quality comparison without repeating ibUMAP CUDA under different populations. It is determined by coverage, not selected by quality outcomes. Runtime and stability comparisons retain the full available pairs: 66 datasets versus umap-learn, 71 versus cuML, and 66 versus TorchDR. The additional five datasets remain in the complete GPU quality summaries below. In the manifest, daggers mark datasets outside the common set.

\paragraph{Inputs and common settings.}
Algorithms receive the same processed feature file for a dataset, verified by its SHA-256 hash and loaded as \texttt{float32}. The stored inputs have dataset-specific preprocessing; the benchmark does not replace these inputs with a common new preprocessing pipeline. Each method builds its own graph and initializes its own embedding through its full fitting path. The common requested parameters are two output dimensions, Euclidean distance, 15 neighbors, 200 epochs, \texttt{min\_dist=0.1}, spread 1, learning rate 1, repulsion strength 1, negative-sample rate 5, fuzzy-set mixing ratio 1, and local connectivity 1. Low-memory mode and \texttt{n\_jobs=-1} are requested where supported. The adapters record effective and unsupported parameters. In particular, TorchDR maps the epoch budget to 200 iterations and does not consume the requested local-connectivity, low-memory, thread-count, or fuzzy-set-mixing parameters. Equal requested budgets do not imply equal update counts, matched convergence, or identical graphs and starts.

\paragraph{ibFFT grid defaults.}
Both ibUMAP paths use interpolation order $p=1$, kernel cap $c_K=4$, and
$\varepsilon=10^{-3}$. At each epoch the mesh spans the square hull of the current
coordinates, and the requested box count per dimension is
$s_B\min\{4\sqrt{n},\,\max[\,2\sqrt{n/\log n},\,B_{\min},\,R/\delta\,]\}$, where $R$ is
the coordinate range, $s_B=1$ is the box-scale multiplier, $B_{\min}=100$, and
$\delta=1$ intervals per unit. The request is then raised to the smallest supported
FFT size strictly above it and capped at $6912$, so the effective minimum is $108$
boxes per dimension. The box width is rounded upward onto a logarithmic scale with
64 levels per octave, which preserves coverage and lets nearby epochs reuse a cached
kernel spectrum. Near-field exactness corrections and kernel sub-sampling are
available in the implementation but disabled in every reported run.

\paragraph{Execution profiles.}
The seeded umap-learn and cuML baselines pass \texttt{random\_state=42}; the unseeded baselines omit it. cuML's graph-routing options and \texttt{force\_serial\_epochs} are left to the installed library defaults. Seeded umap-learn resolves to one job with nonparallel optimization, whereas its unseeded profile permits parallel execution. Seeded ibUMAP sets \texttt{deterministic=true}; its CPU graph policy resolves to eight jobs, and its CUDA deposition mode is segmented. Unseeded ibUMAP disables deterministic mode; the CPU graph policy requests all available jobs, and CUDA deposition is atomic. Both recorded CPU profiles use serial P2M deposition. The seed-cost comparison therefore includes these execution policies, rather than isolating the effect of a random-number seed alone.

TorchDR uses FAISS on CUDA with distributed execution and compilation disabled. Its seeded profile sets seed 42, deterministic PyTorch/cuDNN algorithms, and the deterministic cuBLAS workspace configuration \texttt{:4096:8}. Its unseeded profile disables deterministic algorithms and uses the recorded fast-mode settings. The TorchDR runs use a separate package environment on the same recorded hardware configuration.

\begin{table}[htbp]
  \centering\small
  \caption{Recorded benchmark environment. Values describe the execution host, not the host used to prepare this manuscript.}
  \label{tab:benchmark-environment}
  \begin{tabular}{@{}p{0.29\linewidth}p{0.65\linewidth}@{}}
    \toprule
    Component & Recorded value \\
    \midrule
    GPU & NVIDIA RTX A5000, 24 GiB class \\
    Operating system & Linux 5.15.0-124, x86\_64, glibc 2.31 \\
    Python & 3.11.15 \\
    CPU & Intel Core i7-11700K, 8 cores / 16 hardware threads, 3.60\,GHz nominal frequency \\
    Host RAM & 32\,GB class (\texttt{MemTotal} 31.2\,GiB) \\
    umap-learn / PyNNDescent & 0.5.12 / 0.5.13 \\
    cuML / cuVS / RMM & 26.6.0 / 26.6.0 / 26.6.0 \\
    CuPy / pyFFTW & 13.6.0 / 0.15.1 \\
    NumPy / SciPy / Numba & 1.26.4 / 1.16.3 / 0.61.2 \\
    TorchDR / PyTorch & 0.4 / 2.12.1+cu126 \\
    \bottomrule
  \end{tabular}
\end{table}

The exported hardware signature is shared across the runs and includes the operating-system platform and GPU description. The export records neither CPU model nor host memory; both are supplemented from the execution host, the CPU from \texttt{lscpu} and the memory from \texttt{/proc/meminfo} as captured in the environment record of the downstream experiment (Appendix~\ref{app:downstream-results}), which ran on the same host with the same kernel, Python build, and CPU. The results characterize the recorded execution configuration rather than hardware-independent performance. Per-family runtime and source signatures prevent incompatible results from being silently combined. This appendix uses the accepted frozen bundle, rather than revalidating its numerical claims against a later source checkout.

\paragraph{Timing and repetitions.}
Each run uses a fresh worker process. After loading and validating the input, the worker performs one warmup of its configured algorithm before measuring the fitting call. The measured region includes construction, graph building, initialization, optimization, synchronization, and transfer of the result to host memory. It excludes worker startup, input loading, hashing, warmup, output serialization, evaluation, and plotting. Runs are shuffled within repetition blocks. This is warm-call end-to-end timing, not cold-start latency. The primary field is \texttt{e2e\_call\_wall\_s}; the export's legacy \texttt{optimization\_call\_wall\_s} field is an alias of the same end-to-end measurement and is not analyzed as a separate timing boundary. Resource snapshots do not provide a comparable peak-device-memory measurement, so no such comparison is inferred here.

\paragraph{Aggregation.}
Let $t_{d,a,r}$ denote the recorded runtime on dataset $d$, algorithm/profile $a$, and repetition $r$. We use $\widetilde t_{d,a}=\operatorname{median}_{r=1}^{5}t_{d,a,r}$ and the paired speed ratio $s_d=\widetilde t_{d,\mathrm{baseline}}/\widetilde t_{d,\mathrm{ibUMAP}}$. A ratio above one favors ibUMAP. Suite and size-group summaries take the median of these dataset-level ratios. Main-text fidelity uses five-run mean scores within each dataset, then takes medians across the common 66 datasets. Appendix~\ref{app:benchmark-quality} gives the seeded quality-score convention and distinguishes means, marginal medians, and paired-difference medians. Repeated timings or identical-score reuse records are not treated as additional datasets. Size bins and collection weighting are descriptive summaries of this fixed suite, not a random sample of all potential datasets or a pre-registered noninferiority study.

\FloatBarrier
\subsection{Additional end-to-end performance results}
\label{app:benchmark-runtime}

Table~\ref{tab:benchmark-runtime} separates the CPU fixed-overhead regime from larger datasets and retains both profiles for the GPU comparison. CPU ibUMAP loses on all six datasets below 1,000 observations and wins on every tested larger dataset. The suite has no observations between 646 and 1,567 rows, and the datasets also differ in dimensionality and structure. The first dataset showing a speedup therefore does not establish a controlled CPU crossover.

\begin{table}[htbp]
  \centering\small
  \caption{End-to-end speed ratios, baseline/ibUMAP, summarized by dataset size. S: seeded; U: unseeded. ``Wins'' counts dataset-level ratios strictly greater than one. CPU/UMAP compares ibUMAP CPU with umap-learn; both GPU comparisons use ibUMAP CUDA.}
  \label{tab:benchmark-runtime}
  \setlength{\tabcolsep}{4pt}
  \begin{tabular}{@{}llrrrrr@{}}
    \toprule
    Comparison & Size group & $n$ & Median S & Median U & Wins S & Wins U \\
    \midrule
    CPU/UMAP & All & 66 & 5.792 & 3.286 & 60/66 & 60/66 \\
CPU/UMAP & $N<10^3$ & 6 & 0.310 & 0.167 & 0/6 & 0/6 \\
CPU/UMAP & $10^3\leq N<10^4$ & 13 & 5.724 & 3.832 & 13/13 & 13/13 \\
CPU/UMAP & $10^4\leq N<10^5$ & 41 & 5.937 & 3.042 & 41/41 & 41/41 \\
CPU/UMAP & $10^5\leq N<10^6$ & 5 & 7.310 & 4.077 & 5/5 & 5/5 \\
CPU/UMAP & $N\geq10^6$ & 1 & 5.253 & 4.466 & 1/1 & 1/1 \\
GPU/cuML & All & 71 & 0.726 & 0.875 & 17/71 & 25/71 \\
GPU/cuML & $N<10^4$ & 19 & 0.253 & 0.359 & 2/19 & 3/19 \\
GPU/cuML & $10^4\leq N<10^5$ & 41 & 0.726 & 0.905 & 7/41 & 11/41 \\
GPU/cuML & $10^5\leq N<10^6$ & 5 & 0.990 & 1.062 & 2/5 & 5/5 \\
GPU/cuML & $N\geq10^6$ & 6 & 1.045 & 1.444 & 6/6 & 6/6 \\
GPU/TorchDR & All & 66 & 1.388 & 1.532 & 60/66 & 62/66 \\
 
    \bottomrule
  \end{tabular}
\end{table}

\begin{figure}[htbp]
  \centering
  \includegraphics[width=0.98\linewidth]{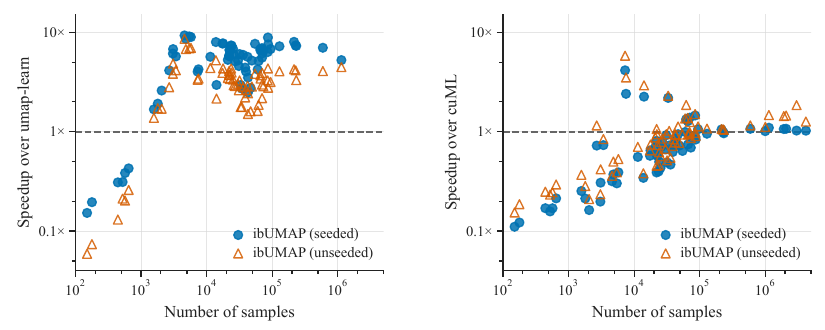}
  \caption{Dataset-level end-to-end speed ratios for CPU (left) and GPU versus cuML (right). Each ratio compares matching seeded or unseeded profiles after taking each algorithm's five-run median time. The horizontal line denotes equal runtime; values above it favor ibUMAP. Both axes are logarithmic. CPU and GPU contain 66 and 71 paired datasets per profile, respectively.}
  \label{fig:appendix-speedups}
\end{figure}

Against cuML, the unseeded median speed ratio rises from $0.359$ below $10^4$ observations to $0.905$ in the $10^4$--$10^5$ bin, $1.062$ in the $10^5$--$10^6$ bin, and $1.444$ at or above $10^6$. Seeded ibUMAP also wins on all six million-scale datasets, but its median advantage is only $1.05\times$: modest, not evidence of a large seeded GPU speedup. Figure~\ref{fig:appendix-speedups} retains all individual paired speed ratios.

\begin{table}[htbp]
  \centering\small
  \caption{Runtime cost and observed output identity of the seeded profile. Time ratios are seeded/unseeded, so values above one indicate a seeded runtime penalty. ``Identical'' counts datasets with identical full embedding hashes in all five seeded runs. The million-scale column has only one dataset for CPU and TorchDR.}
  \label{tab:benchmark-profiles}
  \setlength{\tabcolsep}{5pt}
  \begin{tabular}{@{}lrrrrr@{}}
    \toprule
    Implementation & $n$ & Median ratio & $n_{\geq10^6}$ & Ratio $N\geq10^6$ & Identical \\
    \midrule
    umap-learn & 66 & 1.664 & 1 & 1.170 & 66/66 \\
ibUMAP CPU & 66 & 0.961 & 1 & 0.995 & 66/66 \\
cuML & 71 & 1.071 & 6 & 9.191 & 70/71 \\
ibUMAP CUDA & 71 & 1.306 & 6 & 11.718 & 71/71 \\
TorchDR & 66 & 1.136 & 1 & 1.044 & 66/66 \\
 
    \bottomrule
  \end{tabular}
\end{table}

\begin{figure}[htbp]
  \centering
  \includegraphics[width=\linewidth]{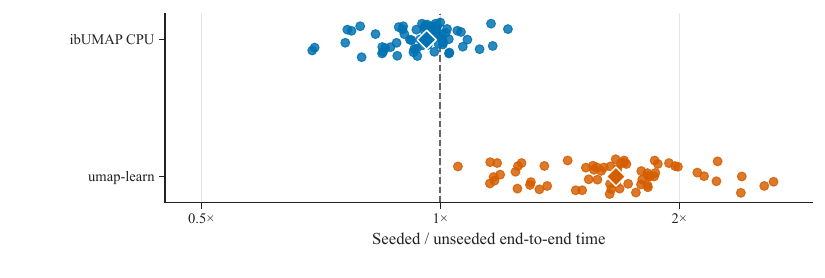}
  \caption{CPU runtime cost of seeded execution. Each point is one dataset's seeded five-run median time divided by its unseeded five-run median, with 66 datasets per implementation. Diamonds mark medians across datasets. The horizontal scale is logarithmic; ratios above one indicate a seeded runtime penalty. Vertical jitter separates points.}
  \label{fig:appendix-cpu-seed-cost}
\end{figure}

The CPU repeatability result concerns the absence of an aggregate runtime penalty: ibUMAP's median ratio is $0.961$, but 4/66 datasets have a seeded median runtime more than 10\% above their unseeded median. Figure~\ref{fig:appendix-cpu-seed-cost} retains all dataset-level ratios underlying the CPU repeatability comparison in Section~\ref{sec:e2e-performance}.

\begin{figure}[htbp]
  \centering
  \includegraphics[width=\linewidth]{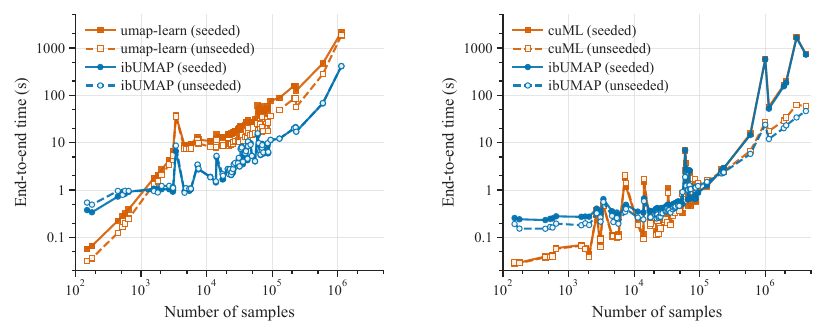}
  \caption{Absolute end-to-end runtime for both execution profiles: CPU versus umap-learn (left, 66 datasets) and GPU versus cuML (right, 71 datasets). Each point is a five-run median. Both axes are logarithmic. Connecting lines are visual guides across different datasets, not controlled size sweeps. Seeded and unseeded fitting paths can have markedly different costs on large GPU workloads.}
  \label{fig:appendix-runtime-profiles}
\end{figure}

\begin{figure}[htbp]
  \centering
  \includegraphics[width=0.49\linewidth]{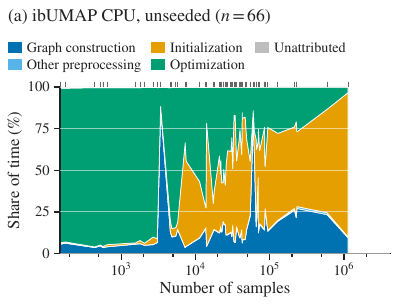}\hfill
  \includegraphics[width=0.49\linewidth]{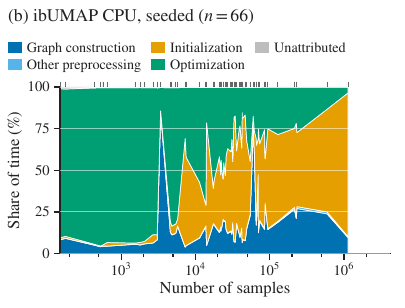}\\[4pt]
  \includegraphics[width=0.49\linewidth]{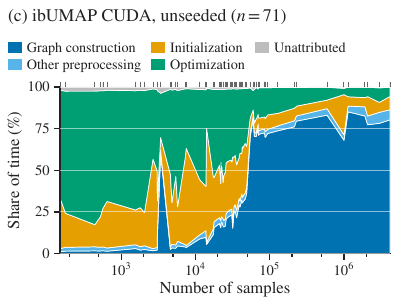}\hfill
  \includegraphics[width=0.49\linewidth]{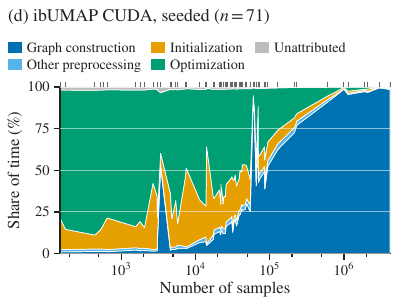}
  \caption{Shares of ibUMAP end-to-end time by stage: CPU (a, b; 66 datasets) and CUDA (c, d; 71 datasets) under both execution profiles. For each dataset, every stage's five-run median time is divided by the sum of these medians. Unattributed time is the measured call time not covered by the recorded stages. Tick marks along the top edge mark dataset sizes. Areas between them are visual interpolations across different datasets, not controlled size sweeps, and the two 70,000-observation datasets are averaged. The horizontal axis is logarithmic and shared by all panels.}
  \label{fig:appendix-stage-share}
\end{figure}

GPU behavior differs substantially. Across the six million-scale datasets, seeded/unseeded ratios have medians $11.718$ for ibUMAP CUDA and $9.191$ for cuML. The ibUMAP ratios range from $4.360$ to $48.215$. Figure~\ref{fig:appendix-runtime-profiles} gives the absolute CPU and GPU runtimes for both profiles. The unseeded GPU speedups in the main text must not be interpreted as the runtime of the seeded configuration. Figure~\ref{fig:appendix-stage-share} attributes this seeded cost almost entirely to graph construction.

The main-text TorchDR comparison uses its 66 datasets with completed unseeded runs. Its seeded counterpart gives ibUMAP CUDA a median speedup of $1.39\times$, with 60/66 wins, compared with $1.53\times$ and 62/66 unseeded. These comparisons do not include the five resource-skipped datasets and are not substituted for the 71-dataset cuML comparison.

\paragraph{Stage composition.}
Figure~\ref{fig:appendix-stage-share} divides each ibUMAP end-to-end time into the stages reported by its internal timers: graph construction, other preprocessing, spectral initialization, optimization, and the unattributed remainder of the measured call. Shares use five-run medians per dataset; absolute times are in Figure~\ref{fig:appendix-runtime-profiles}. Below $10^4$ observations, optimization takes a median 91\% of unseeded CPU time and 67\% of unseeded CUDA time. From $10^4$ observations onward, initialization becomes the largest CPU stage, with median shares of 43\% in the $10^4$--$10^5$ bin and 54\% at or above $10^5$; optimization falls to 22\% in the latter bin. On CUDA, graph construction dominates at or above $10^5$ observations, with median shares of 80\% unseeded and 96\% seeded, while optimization takes 8\% and 2\%. The seeded GPU penalty on the six million-scale datasets therefore arises in graph construction: on Google News, for example, graph construction takes 1,638\,s seeded versus 27\,s unseeded, whereas optimization changes only from 3.0\,s to 3.7\,s. The cause is the neighbor search: GPU NN-Descent does not currently guarantee repeatable graphs, so the seeded profile always builds the $k$-NN graph by brute force, whereas the unseeded profile switches to NN-Descent above 50,000 observations. The isolated peaks in the graph share at smaller sizes correspond to high-dimensional inputs ($D\geq784$, against a suite median of 50). The unattributed remainder has a median share of at most 2\% in every size bin, and no run's recorded stages exceed its measured time. The baselines do not expose comparable stage timers, so this decomposition does not support stage-level comparisons across implementations.

\FloatBarrier
\subsection{Fidelity metrics and additional comparisons}
\label{app:benchmark-quality}

\subsubsection{Evaluation measures and aggregation}

All five fidelity metrics use the processed feature representation used for fitting. Evaluation uses all rows when $n\leq10^6$, otherwise a uniform sample of $10^6$ rows without replacement. All five metrics use sampling seed 42 and the same evaluation indices across methods and runs. Local scores describe neighborhoods of the induced evaluation set. The full embedding is fitted before sampling; identical embedding hashes permit score reuse while preserving each run's provenance.

\paragraph{Local fidelity measures.}
Trustworthiness penalizes embedding-space neighbors that are distant in input-space rank; continuity penalizes input-space neighbors that are distant in embedding-space rank \citep{venna2006local,lee2009quality}. Neighborhood preservation reports the mean fraction of shared neighbors between the input and embedding neighbor sets \citep{huang2022comprehensive}. All use $k=15$, exclude self-neighbors, and are higher-is-better scores.

Let $N_X^k(i)$ and $N_Y^k(i)$ denote the $k$ nearest neighbors of observation $i$ in the input and embedding spaces, and let $r_X(i,j)$ and $r_Y(i,j)$ be the corresponding one-based neighbor ranks. With $U_i=N_Y^k(i)\setminus N_X^k(i)$, $V_i=N_X^k(i)\setminus N_Y^k(i)$, and $Z=2/[nk(2n-3k-1)]$, the scores are
\begin{align}
  \mathrm{TW}(k) &= 1-Z\sum_{i=1}^n\sum_{j\in U_i}\bigl(r_X(i,j)-k\bigr), \\
  \mathrm{C}(k) &= 1-Z\sum_{i=1}^n\sum_{j\in V_i}\bigl(r_Y(i,j)-k\bigr), \\
  \mathrm{NP}(k) &= \frac{1}{nk}\sum_{i=1}^n\left|N_X^k(i)\cap N_Y^k(i)\right|.
\end{align}
TW and C weight rank errors, whereas NP counts retained neighbors without rank weighting. These measures characterize local neighborhood fidelity.

\paragraph{Global fidelity measures.}
RTA is the fraction of sampled triplets $(i,j,k)$ preserving whether $j$ or $k$ is closer to anchor $i$; $\rho_D$ is the Spearman correlation of input and embedding distances over sampled point pairs \citep{huang2022comprehensive}. Both use Euclidean distances on the stored \texttt{float32} arrays. We draw $10^6$ ordered triplets with replacement, with distinct indices within each triplet, exclude input-distance ties, and count embedding-distance ties as incorrect. For $m$ evaluation points, we sample $\min\{10^6,m(m-1)/2\}$ distinct unordered pairs without replacement, using all pairs when fewer are available; Spearman uses average ranks for ties. Triplets and pairs use seed 42 and are shared across methods and runs. RTA lies in $[0,1]$ and $\rho_D$ in $[-1,1]$, with higher values preferred. Unlike the stability measure below, $\rho_D$ compares inputs with embeddings, not two reruns. These global scores cover all nine scored mechanism variants and both end-to-end profiles on their full available dataset sets.

The seeded profile is intended to reproduce the same embedding on the same device, so all five quality metrics use its first run; additional runs measure runtime and check output identity (Appendix~\ref{app:benchmark-stability}). Unseeded quality uses the mean of five runs. For metric $q$, let $\bar q_{d,a}$ denote this dataset-level score for implementation/profile $a$. On a specified dataset set $D$, the absolute mean and marginal median are $\operatorname{mean}_{d\in D}\bar q_{d,a}$ and $\operatorname{median}_{d\in D}\bar q_{d,a}$. The main table uses the latter on the common 66 datasets. A paired difference is instead $\Delta_d=\bar q_{d,\mathrm{ibUMAP}}-\bar q_{d,\mathrm{baseline}}$. Its mean equals the difference of means on a shared set, whereas its median need not equal the difference of marginal medians. Neither summary counts per-dataset wins or establishes statistical significance. No composite quality score or post hoc noninferiority tolerance is introduced.

\subsubsection{Extended fidelity summaries}

Tables~\ref{tab:benchmark-quality-absolute-mean} and~\ref{tab:benchmark-quality-absolute-median} cover both profiles on the common 66 datasets and the full 71-dataset cuML--ibUMAP CUDA pair. Bold marks the best score within each coverage block and profile.

\begin{table}[htbp]
  \centering\small
  \setlength{\tabcolsep}{4pt}
  \caption{Mean fidelity by profile and coverage. Dataset scores use the first seeded run or the five-run unseeded mean; datasets receive equal weight. Bold marks the best score within each block and profile.}
  \label{tab:benchmark-quality-absolute-mean}
  \begin{tabular}{@{}llrrrrr@{}}
    \toprule
    Method & Profile & TW $\uparrow$ & C $\uparrow$ & NP $\uparrow$ & RTA $\uparrow$ & $\rho_D$ $\uparrow$ \\
    \midrule
    \multicolumn{7}{@{}l}{\emph{CPU: common 66 datasets}} \\
umap-learn & Unseeded & \textbf{0.9728} & \textbf{0.985230} & \textbf{0.3983} & \textbf{0.7254} & \textbf{0.5952} \\
ibUMAP & Unseeded & 0.9672 & 0.984801 & 0.3762 & 0.6961 & 0.5292 \\
umap-learn & Seeded & \textbf{0.9727} & \textbf{0.985119} & \textbf{0.3979} & \textbf{0.7286} & \textbf{0.6006} \\
ibUMAP & Seeded & 0.9673 & 0.984823 & 0.3758 & 0.6971 & 0.5306 \\
\midrule
\multicolumn{7}{@{}l}{\emph{GPU: common 66 datasets}} \\
cuML & Unseeded & \textbf{0.9715} & 0.985300 & 0.3927 & 0.7299 & 0.6037 \\
TorchDR & Unseeded & 0.9710 & 0.984591 & \textbf{0.4028} & \textbf{0.7532} & \textbf{0.6435} \\
ibUMAP & Unseeded & 0.9671 & \textbf{0.985545} & 0.3758 & 0.7139 & 0.5640 \\
cuML & Seeded & 0.9680 & 0.985258 & 0.3812 & 0.7263 & 0.5939 \\
TorchDR & Seeded & \textbf{0.9712} & 0.984450 & \textbf{0.4030} & \textbf{0.7529} & \textbf{0.6424} \\
ibUMAP & Seeded & 0.9671 & \textbf{0.985612} & 0.3761 & 0.7108 & 0.5553 \\
\midrule
\multicolumn{7}{@{}l}{\emph{GPU: full 71 datasets}} \\
cuML & Unseeded & \textbf{0.9652} & \textbf{0.982812} & \textbf{0.3664} & \textbf{0.7243} & \textbf{0.5912} \\
ibUMAP & Unseeded & 0.9602 & 0.982737 & 0.3506 & 0.7080 & 0.5504 \\
cuML & Seeded & \textbf{0.9623} & 0.983954 & \textbf{0.3557} & \textbf{0.7202} & \textbf{0.5795} \\
ibUMAP & Seeded & 0.9605 & \textbf{0.984558} & 0.3509 & 0.7054 & 0.5422 \\
 
    \bottomrule
  \end{tabular}
\end{table}

\begin{table}[htbp]
  \centering\small
  \setlength{\tabcolsep}{4pt}
  \caption{Median fidelity for the same blocks and profiles as Table~\ref{tab:benchmark-quality-absolute-mean}, using the first seeded run or the five-run unseeded mean per dataset. Common-set unseeded rows reproduce Table~\ref{tab:e2e-fidelity-common} with extra precision for C. Bold marks the best score before rounding.}
  \label{tab:benchmark-quality-absolute-median}
  \begin{tabular}{@{}llrrrrr@{}}
    \toprule
    Method & Profile & TW $\uparrow$ & C $\uparrow$ & NP $\uparrow$ & RTA $\uparrow$ & $\rho_D$ $\uparrow$ \\
    \midrule
    \multicolumn{7}{@{}l}{\emph{CPU: common 66 datasets}} \\
umap-learn & Unseeded & \textbf{0.9901} & 0.994200 & \textbf{0.3699} & \textbf{0.7249} & \textbf{0.6072} \\
ibUMAP & Unseeded & 0.9853 & \textbf{0.995031} & 0.3436 & 0.7023 & 0.5493 \\
umap-learn & Seeded & \textbf{0.9902} & 0.994569 & \textbf{0.3677} & \textbf{0.7259} & \textbf{0.6031} \\
ibUMAP & Seeded & 0.9852 & \textbf{0.995017} & 0.3443 & 0.6995 & 0.5396 \\
\midrule
\multicolumn{7}{@{}l}{\emph{GPU: common 66 datasets}} \\
cuML & Unseeded & \textbf{0.9892} & 0.994996 & 0.3635 & 0.7349 & 0.6295 \\
TorchDR & Unseeded & 0.9873 & 0.994347 & \textbf{0.3768} & \textbf{0.7647} & \textbf{0.6706} \\
ibUMAP & Unseeded & 0.9851 & \textbf{0.995005} & 0.3407 & 0.7217 & 0.5897 \\
cuML & Seeded & 0.9864 & \textbf{0.995175} & 0.3502 & 0.7288 & 0.6076 \\
TorchDR & Seeded & \textbf{0.9873} & 0.994281 & \textbf{0.3776} & \textbf{0.7648} & \textbf{0.6692} \\
ibUMAP & Seeded & 0.9852 & 0.995017 & 0.3413 & 0.7179 & 0.5789 \\
\midrule
\multicolumn{7}{@{}l}{\emph{GPU: full 71 datasets}} \\
cuML & Unseeded & \textbf{0.9878} & \textbf{0.993971} & \textbf{0.3414} & \textbf{0.7266} & \textbf{0.6061} \\
ibUMAP & Unseeded & 0.9841 & 0.993892 & 0.3267 & 0.7156 & 0.5725 \\
cuML & Seeded & \textbf{0.9864} & \textbf{0.994454} & \textbf{0.3349} & \textbf{0.7221} & \textbf{0.5939} \\
ibUMAP & Seeded & 0.9844 & 0.993905 & 0.3329 & 0.7140 & 0.5500 \\
 
    \bottomrule
  \end{tabular}
\end{table}

Table~\ref{tab:benchmark-quality} summarizes differences within each baseline pair using its full available dataset set (66/71/66). It retains both the mean and the median of paired differences, separately from the marginal score medians above.

\begin{table}[htbp]
  \centering\small
  \caption{Means and medians of paired fidelity differences, ibUMAP minus baseline, on each pair's full available set. Dataset-level scores are paired before aggregation: the first seeded run or the mean of five unseeded runs. Positive differences favor ibUMAP for all five metrics. C retains six decimals in the median panel to avoid rounding a small negative difference to zero.}
  \label{tab:benchmark-quality}
  \setlength{\tabcolsep}{3.5pt}
  \begin{tabular}{@{}llrrrrrr@{}}
    \toprule
    Comparison & Profile & $n$ & $\Delta$TW & $\Delta$C & $\Delta$NP & $\Delta$RTA & $\Delta\rho_D$ \\
    \midrule
    \multicolumn{8}{@{}l}{\emph{Mean of paired differences}} \\
    CPU/UMAP & Seeded & 66 & $-0.0054$ & $-0.0003$ & $-0.0221$ & $-0.0315$ & $-0.0700$ \\
CPU/UMAP & Unseeded & 66 & $-0.0056$ & $-0.0004$ & $-0.0221$ & $-0.0293$ & $-0.0660$ \\
GPU/cuML & Seeded & 71 & $-0.0018$ & $+0.0006$ & $-0.0048$ & $-0.0148$ & $-0.0372$ \\
GPU/cuML & Unseeded & 71 & $-0.0051$ & $-0.0001$ & $-0.0158$ & $-0.0162$ & $-0.0408$ \\
GPU/TorchDR & Seeded & 66 & $-0.0041$ & $+0.0012$ & $-0.0269$ & $-0.0421$ & $-0.0870$ \\
GPU/TorchDR & Unseeded & 66 & $-0.0039$ & $+0.0010$ & $-0.0270$ & $-0.0393$ & $-0.0795$ \\
 
    \midrule
    \multicolumn{8}{@{}l}{\emph{Median of paired differences}} \\
    CPU/UMAP & Seeded & 66 & $-0.0028$ & $+0.000094$ & $-0.0213$ & $-0.0191$ & $-0.0541$ \\
CPU/UMAP & Unseeded & 66 & $-0.0030$ & $-0.000034$ & $-0.0222$ & $-0.0155$ & $-0.0522$ \\
GPU/cuML & Seeded & 71 & $-0.0010$ & $-0.000126$ & $-0.0038$ & $-0.0049$ & $-0.0210$ \\
GPU/cuML & Unseeded & 71 & $-0.0031$ & $-0.000051$ & $-0.0163$ & $-0.0106$ & $-0.0257$ \\
GPU/TorchDR & Seeded & 66 & $-0.0025$ & $+0.000483$ & $-0.0304$ & $-0.0307$ & $-0.0549$ \\
GPU/TorchDR & Unseeded & 66 & $-0.0021$ & $+0.000587$ & $-0.0297$ & $-0.0303$ & $-0.0532$ \\
 
    \bottomrule
  \end{tabular}
\end{table}

\FloatBarrier
\subsubsection{Dataset-level variation and aggregation sensitivity}

\begin{figure}[htbp]
  \centering
  \includegraphics[width=\linewidth]{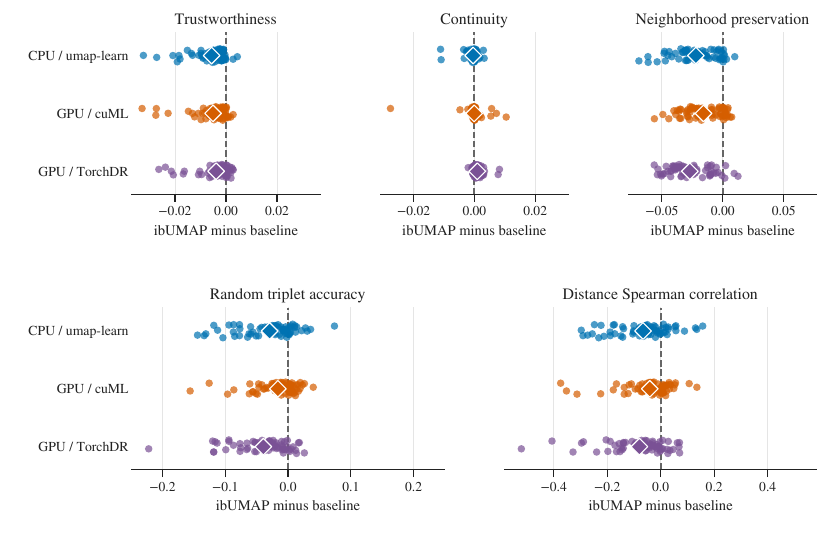}
  \caption{Unseeded fidelity differences on the full available pairs: CPU/umap-learn (66 datasets), GPU/cuML (71), and GPU/TorchDR (66). Each point is ibUMAP's five-run mean score minus the baseline's; diamonds mark means across datasets. Positive differences favor ibUMAP. Horizontal scales differ between metrics and are shared across comparisons within each panel. Vertical jitter separates overlapping points.}
  \label{fig:e2e-fidelity}
\end{figure}

Figure~\ref{fig:e2e-fidelity} retains all individual unseeded differences in the five fidelity metrics on the full paired sets. Unlike the common-set main table, its cuML comparison includes all 71 datasets. The spread shows why a single mean or median cannot characterize every dataset.

Summary choices can change the apparent ordering. On the common 66 datasets, CPU ibUMAP has a higher continuity marginal median than umap-learn, but its mean difference is $-0.000429$ and its paired-difference median is $-0.000034$. The higher marginal median therefore does not demonstrate an improvement on a typical paired dataset. Reporting both summaries exposes this change instead of treating the most favorable summary as a uniform quality advantage. Small continuity differences also do not cancel the larger changes in neighborhood preservation.

\paragraph{Collection weighting.}
The suite contains 35 Tabula Sapiens entries, 11 C.~elegans entries, and 4 MERFISH entries, alongside 21 other datasets. Several entries represent related or overlapping subsets. As a post hoc aggregation check, we average the paired score differences within each of these three collections, retain each remaining dataset as a singleton, and then weight the resulting blocks equally. This gives 24 blocks for the cuML comparison and 22 for CPU and TorchDR after resource skips. It reduces the multiplicity of these three collections; it does not establish independent biological replicates or a uniquely correct population weighting.

\begin{table}[htbp]
  \centering\small
  \caption{Collection-weighted mean fidelity differences on the full paired sets, using the explicit grouping above. Positive differences favor ibUMAP. The corresponding dataset-weighted means are in the mean panel of Table~\ref{tab:benchmark-quality}.}
  \label{tab:benchmark-collection-quality}
  \begin{tabular}{@{}llrrrrrr@{}}
    \toprule
    Comparison & Profile & Blocks & $\Delta$TW & $\Delta$C & $\Delta$NP & $\Delta$RTA & $\Delta\rho_D$ \\
    \midrule
    CPU/UMAP & Seeded & 22 & $-0.0029$ & $-0.0011$ & $-0.0083$ & $-0.0361$ & $-0.0717$ \\
CPU/UMAP & Unseeded & 22 & $-0.0029$ & $-0.0016$ & $-0.0077$ & $-0.0290$ & $-0.0608$ \\
GPU/cuML & Seeded & 24 & $-0.0015$ & $+0.0021$ & $+0.0012$ & $-0.0014$ & $+0.0024$ \\
GPU/cuML & Unseeded & 24 & $-0.0044$ & $-0.0002$ & $-0.0041$ & $-0.0092$ & $-0.0192$ \\
GPU/TorchDR & Seeded & 22 & $-0.0071$ & $+0.0023$ & $-0.0195$ & $-0.0268$ & $-0.0496$ \\
GPU/TorchDR & Unseeded & 22 & $-0.0068$ & $+0.0017$ & $-0.0195$ & $-0.0274$ & $-0.0500$ \\
 
    \bottomrule
  \end{tabular}
\end{table}

All three full-coverage unseeded comparisons retain negative mean differences in TW, NP, RTA, and $\rho_D$ after this reweighting. Their magnitudes depend on the weighting: for example, the trustworthiness deficit versus TorchDR increases ($-0.0039$ to $-0.0068$). Weighting can change other conclusions: seeded GPU differences in NP and $\rho_D$ versus cuML become slightly positive. We therefore keep the grouping explicit and retain the complete dataset-level inputs instead of selecting datasets by the size or direction of their observed effects.

\FloatBarrier
\subsection{Repeatability and run-to-run stability protocol}
\label{app:benchmark-stability}

\paragraph{Seeded output identity.}
For each dataset and implementation, the five complete saved embedding files are compared by SHA-256 hash. All seeded ibUMAP CPU, ibUMAP CUDA, umap-learn, and TorchDR groups are identical. Seeded cuML is identical on 70/71 datasets. On Google News, its first run differs from runs 2--5, which are identical to one another, despite using the same device, configuration, and \texttt{random\_state=42}. The embedding arrays and their local quality scores differ, confirming that the discrepancy is not merely in file serialization. Thus, setting cuML's random seed did not ensure bitwise repeatability in every tested case under the recorded library defaults. The benchmark rebuilds the graph and initialization on each call, so this concerns full-pipeline repeatability. No unseeded dataset/implementation group has identical hashes across all five runs.

\paragraph{Geometric comparisons.}
For each dataset, implementation, and profile, repetition 1 is compared with repetitions 2--5, giving four comparisons per group, rather than all ten unordered pairs. A deterministic comparison-specific seed selects at most 10,000 point indices without replacement; corresponding points are shared between the two embeddings in that comparison. Different algorithms and repetition pairs can use different sampled indices. Local stability is the mean fraction of shared neighbors between their 15-nearest-neighbor sets within this sampled point set.

For geometric alignment, each sampled embedding is centered and divided by its Frobenius norm. An orthogonal Procrustes fit, allowing reflection, aligns the coordinates before RMS displacement is computed. The distance-correlation measure uses randomly drawn index pairs, up to $\min(20{,}000,4m)$ for $m$ sampled points, discarding self-pairs. It reports Spearman correlation between distances in the two embeddings. These similarity measures are not affected by global translations, rotations, or uniform scale changes in the way raw coordinate differences would be. The Procrustes RMS depends on this normalization and should not be read as a displacement in the original embedding's units.

Each dataset's score is the median over its four reference-to-repeat comparisons. Table~\ref{tab:benchmark-stability} then takes medians across datasets. Table~\ref{tab:benchmark-stability-pairs} compares algorithms within each shared dataset before counting improvements and aggregating differences; this preserves the appropriate population for each baseline. Figure~\ref{fig:e2e-results}c displays the 15-NN overlap differences with their cross-dataset medians as diamonds, matching the paired medians in the table.

\begin{table}[htbp]
  \centering\small
  \caption{Unseeded run-to-run stability. Each entry is a median across dataset-level median scores. Higher neighbor overlap and distance correlation, and lower aligned RMS, indicate greater stability.}
  \label{tab:benchmark-stability}
  \begin{tabular}{@{}lrrrr@{}}
    \toprule
    Implementation & $n$ & Neighbor overlap & Distance correlation & Procrustes RMS \\
    \midrule
    umap-learn & 66 & 0.6929 & 0.9454 & 0.002100 \\
ibUMAP CPU & 66 & 0.8369 & 0.9902 & 0.000753 \\
cuML & 71 & 0.7311 & 0.9826 & 0.001065 \\
ibUMAP CUDA & 71 & 0.8851 & 0.9983 & 0.000337 \\
TorchDR & 66 & 0.6851 & 0.9632 & 0.001345 \\
 
    \bottomrule
  \end{tabular}
\end{table}

\begin{table}[htbp]
  \centering\small
  \caption{Dataset-paired unseeded stability comparisons. Differences are ibUMAP minus baseline. A difference of medians in Table~\ref{tab:benchmark-stability} need not equal the median of paired differences shown here.}
  \label{tab:benchmark-stability-pairs}
  \setlength{\tabcolsep}{4pt}
  \begin{tabular}{@{}lrrrrr@{}}
    \toprule
    Comparison & $n$ & Overlap wins & Median $\Delta$overlap & Corr.\ wins & Median $\Delta$corr. \\
    \midrule
    CPU/UMAP & 66 & 58/66 & $+0.1436$ & 52/66 & $+0.0357$ \\
GPU/cuML & 71 & 67/71 & $+0.1308$ & 68/71 & $+0.0147$ \\
GPU/TorchDR & 66 & 66/66 & $+0.1961$ & 62/66 & $+0.0256$ \\
 
    \bottomrule
  \end{tabular}
\end{table}

ibUMAP improves neighbor overlap on 58/66 datasets versus umap-learn, 67/71 versus cuML, and 66/66 versus TorchDR. Distance correlation improves on 52/66, 68/71, and 62/66, respectively. The advantage is therefore prevalent rather than universal. It describes variation across unseeded full-pipeline reruns, which can differ in graph construction, initialization, and execution order; it is not a controlled study of initialization perturbations or a causal attribution to the repulsive-field evaluator. Lower run-to-run variation also does not imply higher fidelity to the input data.

\FloatBarrier
\subsection{Complete dataset manifest}

Table~\ref{tab:benchmark-datasets} preserves the complete suite, including small datasets and related subsets. TS abbreviates the \texttt{tabula\_sapiens\_v2} prefix; CE abbreviates \texttt{c\_elegans\_embryogenesis}; MERFISH abbreviates \texttt{whole\_mouse\_brain\_merfish}. Spaces replace underscores in the displayed identifiers, and a slash separates paired cell-type names. The scDEED prefix denotes dataset provenance, not an additional scDEED score in this benchmark. $D$ is the number of features in the processed fitting input.

\paragraph{Data sources.} Tabula Sapiens v2 \citep{tabulasapiens2026atlas}; \textit{C.~elegans} embryogenesis \citep{packer2019lineage} in the CaltechDATA AnnData release \citep{daveigabeltrame2021packer}; whole-mouse-brain MERFISH \citep{zhang2023molecularly}; the scDEED datasets as released by \citet{xia2024scdeed}; MNIST \citep{lecun1998gradient}, Fashion-MNIST \citep{xiao2017fashion}, and CIFAR-10 \citep{krizhevsky2009learning}; GIST-960 \citep{jegou2011product} from ANN-Benchmarks \citep{aumuller2020ann}; Google News word2vec vectors \citep{mikolov2013distributed}; bank, epileptic, HIVA, SECOM, seismic, and spambase as preprocessed by \citet{espadoto2021toward} from their original sources \citep{moro2014data,andrzejak2001indications,wu2017epileptic,guyon2007agnostic,mccann2008secom,sikora2010seismic,hopkins1999spambase}; forest type \citep{johnson2012forest}; and iris, wine, breast cancer, digits, and diabetes via scikit-learn \citep{pedregosa2011scikit,fisher1936use,aeberhard1991wine,street1993nuclear,alpaydin1998optdigits,efron2004least}.

{\small
\setlength{\LTcapwidth}{\linewidth}
\renewcommand{\arraystretch}{0.95}
\begin{longtable}{@{}>{\raggedright\arraybackslash}p{0.70\linewidth}rr@{}}
  \caption{All 71 benchmark datasets and their processed shapes. The 66 unmarked entries form the common quality set in Table~\ref{tab:e2e-fidelity-common}. A dagger marks the five additional datasets evaluated only with cuML and ibUMAP CUDA; both profiles of the other implementations are resource-skipped.}\label{tab:benchmark-datasets}\\
  \toprule
  Dataset & $N$ & $D$ \\
  \midrule
  \endfirsthead
  \multicolumn{3}{l}{\tablename~\thetable{} (continued)}\\
  \toprule
  Dataset & $N$ & $D$ \\
  \midrule
  \endhead
  \midrule
  \multicolumn{3}{r}{Continued on next page}\\
  \endfoot
  \bottomrule
  \endlastfoot
  iris & 150 & 4 \\
wine & 178 & 13 \\
diabetes & 442 & 10 \\
foresttype & 523 & 27 \\
breast cancer & 569 & 30 \\
seismic & 646 & 24 \\
secom & 1,567 & 590 \\
digits & 1,797 & 64 \\
bank & 2,059 & 63 \\
TS: neural & 2,685 & 50 \\
TS: ear & 3,055 & 50 \\
hiva & 3,076 & 1,617 \\
scDEED: velocity & 3,396 & 25,919 \\
spambase & 4,601 & 57 \\
scDEED: marrow & 4,821 & 50 \\
CE: stage early annotated & 5,442 & 32 \\
epileptic & 5,750 & 178 \\
TS: germline & 7,276 & 50 \\
TS: testis & 7,513 & 50 \\
TS: kidney & 11,376 & 50 \\
CE: pair hypodermis / ciliated amphid neuron & 13,836 & 32 \\
TS: pancreas & 14,140 & 50 \\
CE: stage late annotated & 17,436 & 32 \\
TS: skin & 17,786 & 50 \\
TS: prostate & 21,030 & 50 \\
CE: pair body wall muscle / ciliated non amphid neuron & 21,988 & 32 \\
TS: uterus & 22,029 & 50 \\
TS: liver & 22,214 & 50 \\
TS: trachea & 22,671 & 50 \\
CE: pair body wall muscle / ciliated amphid neuron & 23,610 & 32 \\
CE: plot cell type top 8 mixture & 24,095 & 32 \\
CE: pair body wall muscle / hypodermis & 25,266 & 32 \\
TS: heart & 25,832 & 50 \\
TS: bone marrow & 27,112 & 50 \\
TS: large intestine & 30,084 & 50 \\
TS: mammary & 30,936 & 50 \\
CE: stage mid annotated & 31,771 & 32 \\
TS: stomach & 33,064 & 50 \\
TS: eye & 34,273 & 50 \\
CE: cell type top 4 mixture & 35,824 & 32 \\
TS: tongue & 38,754 & 50 \\
TS: salivary gland & 39,821 & 50 \\
TS: small intestine & 42,036 & 50 \\
TS: vasculature & 42,650 & 50 \\
TS: thymus & 42,729 & 50 \\
TS: muscle & 46,772 & 50 \\
TS: ovary & 48,951 & 50 \\
CE: global qc annotated & 54,649 & 32 \\
cifar10 & 60,000 & 3,072 \\
scDEED: cart & 62,167 & 1,000 \\
TS: lung & 65,847 & 50 \\
TS: bladder & 66,385 & 50 \\
fashion mnist & 70,000 & 784 \\
mnist784 & 70,000 & 784 \\
TS: spleen & 70,448 & 50 \\
TS: endothelium & 73,195 & 50 \\
TS: blood & 85,233 & 50 \\
CE: qc passed & 86,024 & 32 \\
scDEED: samusik & 86,864 & 38 \\
TS: fat & 94,415 & 50 \\
TS: lymph node & 129,062 & 50 \\
TS: epithelium & 228,032 & 50 \\
TS: stromal & 232,684 & 50 \\
TS: immune & 592,317 & 50 \\
TS: all cells & 1,136,218 & 50 \\
MERFISH: animal4 sagittal & 215,278 & 50 \\
gist 960 euclidean$^{\dagger}$ & 1,000,000 & 960 \\
MERFISH: animal2 coronal$^{\dagger}$ & 1,915,592 & 50 \\
MERFISH: animal3 sagittal$^{\dagger}$ & 2,081,549 & 50 \\
google news 300d$^{\dagger}$ & 3,000,000 & 300 \\
MERFISH: animal1 coronal$^{\dagger}$ & 4,167,869 & 50 \\
 
\end{longtable}
}

\section{Downstream Reuse: BRAQUE Protocol and Additional Results}
\label{app:downstream-results}

\paragraph{Data and preprocessing.}
The case uses the public L2 single-cell protein-expression sample associated with BRAQUE \citep{dallolio2023braque}. A public antibody workbook supplies a proxy for the paper-specific feature-selection reference: we retain unambiguous normalized name matches with a nonmissing significance entry that does not contain \texttt{IGNORE}. This yields 62 features for 56,962 cells. Lognormal Shrinkage is computed once; its output is median-centered and divided by each feature's mean absolute deviation about its mean, then stored as \texttt{float32}. All repetitions use the same frozen matrix and cell order. The experiment has neither verified spatial coordinates nor final expert phenotype labels, so it evaluates computational partitions rather than spatial organization, cell-type accuracy, or expert annotation quality.

\paragraph{Embedding and clustering configurations.}
Both embedding methods use Euclidean distance, two dimensions, 50 neighbors, 200 epochs, spectral initialization, \texttt{min\_dist=0}, spread 1, learning rate 1, repulsion strength 1, negative-sample rate 5, local connectivity 1, fuzzy-set mixing ratio 1, and low-memory mode. Each fit rebuilds the graph and initialization from the frozen matrix; graphs and starting coordinates are not shared between fits. Seeded profiles set \texttt{random\_state=42}; seeded ibUMAP additionally enables deterministic execution. Seeded UMAP uses one job, while unseeded UMAP and both ibUMAP profiles request eight. The process-level Numba and BLAS thread budget is eight. These are execution-profile comparisons, not an isolation of the random-number seed or optimizer alone.

HDBSCAN uses the raw fitted coordinates, Euclidean distance, \texttt{min\_cluster\_size=10}, \texttt{min\_samples=10}, \texttt{cluster\_selection\_epsilon=0.1}, excess-of-mass selection, the \texttt{best} algorithm, approximate minimum-spanning-tree construction, and one core-distance job. This configuration is identical in all 20 runs. Rigid alignment is applied only for display in Figure~\ref{fig:braque-repeatability}, allowing rotation/reflection and translation but no scaling.

\paragraph{Execution and timing.}
The experiment runs on an Intel Core i7-11700K at 3.60\,GHz with 32\,GB-class RAM (\texttt{MemTotal} 31.2\,GiB), Linux 5.15, Python 3.11.15, umap-learn 0.5.12, HDBSCAN 0.8.44, NumPy 1.26.4, SciPy 1.16.3, scikit-learn 1.8.0, Numba 0.61.2, PyNNDescent 0.5.13, and pyFFTW 0.15.1. Each run uses a fresh process. Four profiles are randomly interleaved within each repetition block. Within each process, embedding and clustering are warmed up on the same fixed 5,000-cell subset, after which fresh estimators fit the complete matrix. All five measured runs per profile are retained; all 20 scheduled runs succeed. The warmup does not guarantee removal of every size-dependent first-call overhead.

Timing separately surrounds \texttt{fit\_transform} and \texttt{fit\_predict}. Embedding time includes graph construction, initialization, and optimization, but excludes estimator construction, input loading, imports, preprocessing, warmup, serialization, and evaluation. The combined time is the sum of the two fit calls within a run, summarized by its five-run median; it is not the complete BRAQUE runtime. These boundaries differ from Section~\ref{sec:scaling}, which also times estimator construction and host materialization.

\paragraph{Evaluation and aggregation.}
Clustering metrics use all cells. ARI treats noise as a separate label. Assignment disagreement uses Hungarian one-to-one matching to maximize non-noise cell overlap, reserves the noise label, and gives unmatched clusters new labels. It therefore includes cluster splits, merges, and noise transitions rather than measuring error against ground truth. We summarize all ten unordered pairs within each profile by their median; these pairs share runs and are not ten independent replicates. Full coordinate and raw-label hashes test exact within-profile repeatability.

Distance correlation uses 200,000 sampled non-self cell pairs from the full fitted coordinates. Trustworthiness uses a fixed 3,000-cell induced subset; input-neighbor overlap uses a fixed 2,048-cell induced subset. Both local metrics use 15 neighbors. The same evaluation indices are used for every fit. These local scores concern subset neighborhoods, not full-data neighborhoods, and are distinct from run-to-run agreement. No noninferiority threshold or significance test is applied.

\begin{table}[htbp]
  \centering\small
  \caption{Complete CPU reuse summary. Times are five-run medians in seconds. Distance correlation $\rho$, ARI, and changed-assignment fraction summarize ten unordered run pairs. ``Exact'' denotes identical full coordinates and raw labels across all five runs within that profile.}
  \label{tab:braque-reuse}
  \setlength{\tabcolsep}{4pt}
  \begin{tabular}{@{}lrrrrrc@{}}
    \toprule
    Profile & Embedding & + HDBSCAN & $\rho$ & ARI & Changed & Exact \\
    \midrule
    UMAP unseeded & 15.436 & 16.076 & 0.937 & 0.611 & 20.80\% & No \\
UMAP seeded & 49.062 & 49.696 & 1.000 & 1.000 & 0.00\% & Yes \\
ibUMAP unseeded & 7.095 & 7.722 & 0.999 & 0.601 & 22.01\% & No \\
ibUMAP seeded & 7.099 & 7.736 & 1.000 & 1.000 & 0.00\% & Yes \\
 
    \bottomrule
  \end{tabular}
\end{table}
\FloatBarrier

For the two plotted runs, 16.29\% of cells change between non-noise clusters and 3.49\% change noise status, totaling 19.78\%; ARI restricted to cells assigned to clusters in both runs is $0.656$. The 20.80\% in Table~\ref{tab:braque-reuse} is the median over all ten UMAP pairs, not the plotted pair's value. Seeded trustworthiness is $0.970230$ for UMAP and $0.962791$ for ibUMAP, a difference of $-0.007440$; input-neighbor overlap is $0.402051$ and $0.370703$, a difference of $-0.031348$. The measured benefit is lower-cost within-profile repeatability, with these fidelity differences retained. It does not imply matching partitions across methods, improved unseeded clustering stability, cross-device reproducibility, or robustness to other seeds.

\begin{figure}[htbp]
  \centering
  \includegraphics[width=\linewidth]{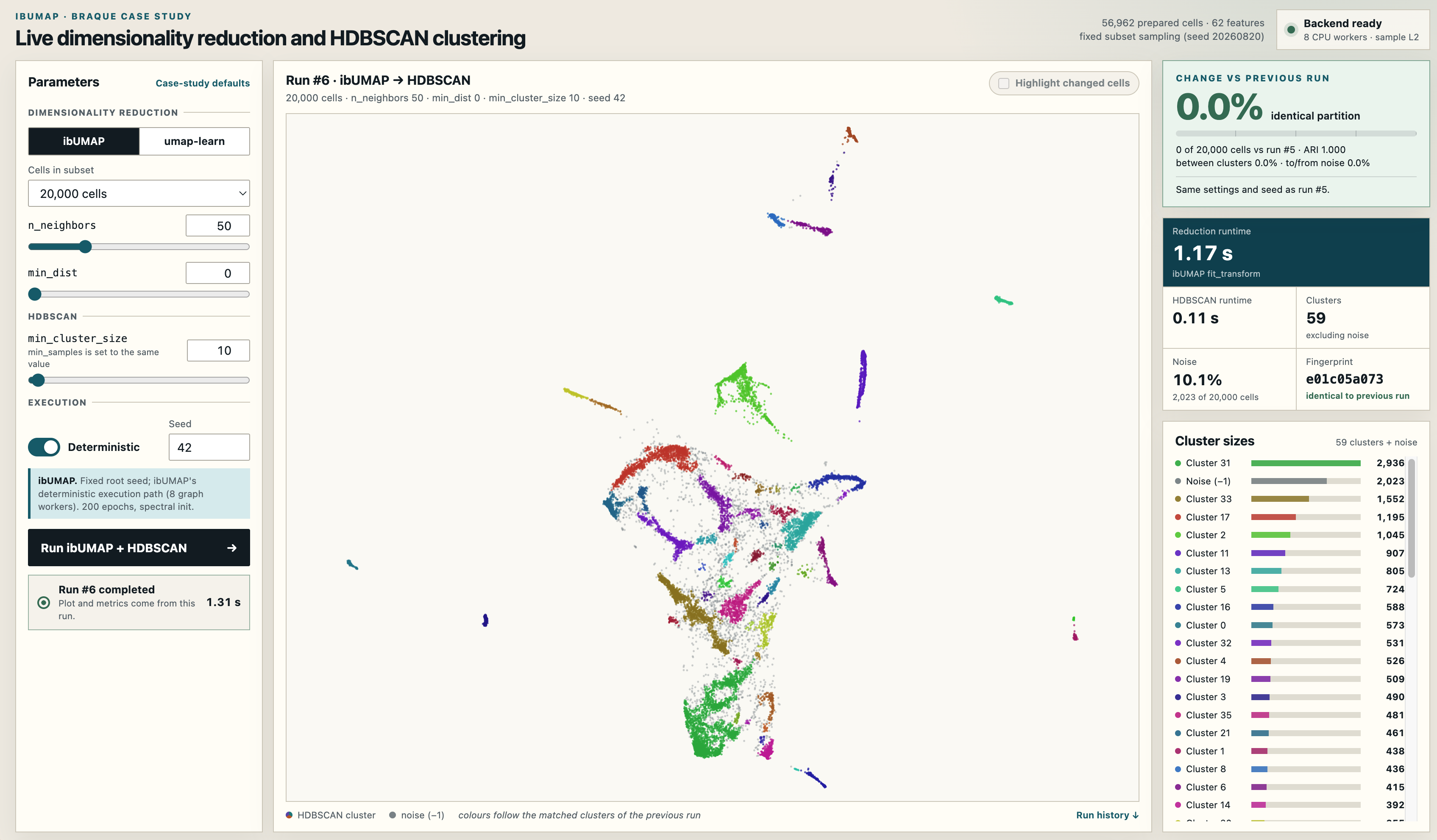}
  \caption{Live demo interface after a repeated seeded ibUMAP run on a 20,000-cell subset. Left: parameters and execution mode. Center: embedding colored by HDBSCAN cluster, with noise in gray. Right: comparison with the previous run (here 0.0\% changed assignments and an identical fingerprint), runtimes, cluster count, noise fraction, and cluster sizes.}
  \label{fig:braque-demo}
\end{figure}

\paragraph{Live demo and video.}
\label{app:live-demo}
The repository also contains a live demo of this workflow (\nolinkurl{experiments/braque/demo_live}, served locally by \nolinkurl{experiments/braque/scripts/09_serve_live_demo.py}; Figure~\ref{fig:braque-demo}). On each request, it runs a fresh ibUMAP or umap-learn fit followed by HDBSCAN on a fixed subset of the frozen L2 matrix; subsets of 1,000 to 20,000 cells are prefixes of one fixed permutation. The adjustable parameters default to the case-study values: \texttt{n\_neighbors}, \texttt{min\_dist}, \texttt{min\_cluster\_size} (with \texttt{min\_samples} set equal), and deterministic mode with a seed. The number of epochs, spectral initialization, and the remaining HDBSCAN settings stay fixed. As in the case study, seeded umap-learn uses one job, whereas seeded ibUMAP runs in deterministic mode with the same number of jobs as its unseeded runs. After each run, the interface reports embedding and clustering times, the cluster count, the noise fraction, and a SHA-256 fingerprint of the coordinates and labels. From the second run on, it also reports the changed-assignment fraction relative to the previous completed run, computed with the matching used for Table~\ref{tab:braque-reuse}.

The accompanying video, linked from the repository, summarizes the method and main results and ends with a screen recording of the demo on a 2,500-cell subset. The recording shows unseeded umap-learn reruns, seeded runs of both methods, and a change of \texttt{n\_neighbors} followed by its reversal, after which the fingerprint equals that of the earlier seeded ibUMAP runs. The demo and the recording illustrate the protocol above. Their timings and change fractions come from single interactive runs on cell subsets and on a different machine, and are not among the reported results.

\FloatBarrier

\end{document}